\documentclass[journal]{IEEEtran}
\usepackage{subfigure}
\usepackage{graphicx}
\usepackage{url}
\usepackage{color}
\usepackage{booktabs}
\usepackage{soul}
\soulregister\cite7
\soulregister\ref7

\ifCLASSINFOpdf
\else
\fi

\usepackage{cite}

\begin{document}

\title{A Novel CNN-based Method for Accurate Ship Detection in HR Optical Remote Sensing Images via Rotated Bounding Box}

\author{Linhao~Li,
        Zhiqiang~Zhou,
        Bo~Wang,
        Lingjuan~Miao
        and~Hua~Zong}% <-this % stops a space

\maketitle

\begin{abstract}
Currently, reliable and accurate ship detection in optical remote sensing images is still challenging. Even the state-of-the-art convolutional neural network (CNN) based methods cannot obtain very satisfactory results. To more accurately locate the ships in diverse orientations, some recent methods conduct the detection via the rotated bounding box. However, it further increases the difficulty of detection, because an additional variable of ship orientation must be accurately predicted in the algorithm. In this paper, a novel CNN-based ship detection method is proposed, by overcoming some common deficiencies of current CNN-based methods in ship detection. Specifically, to generate rotated region proposals, current methods have to predefine multi-oriented anchors, and predict all unknown variables together in one regression process, limiting the quality of overall prediction. By contrast, we are able to predict the orientation and other variables independently, and yet more effectively, with a novel dual-branch regression network, based on the observation that the ship targets are nearly rotation-invariant in remote sensing images. Next, a shape-adaptive pooling method is proposed, to overcome the limitation of typical regular ROI-pooling in extracting the features of the ships with various aspect ratios. Furthermore, we propose to incorporate multilevel features via the spatially-variant adaptive pooling. This novel approach, called multilevel adaptive pooling, leads to a compact feature representation more qualified for the simultaneous ship classification and localization. Finally, detailed ablation study performed on the proposed approaches is provided, along with some useful insights. Experimental results demonstrate the great superiority of the proposed method in ship detection. Code is publicly available at https://github.com/lilinhao/ShipDetection.
\end{abstract}


\IEEEpeerreviewmaketitle

\section{Introduction} \label{section:Introduction}

\IEEEPARstart{T}{he} technique of automatic ship detection in optical remote sensing images can find important application in various tasks, ranging from maritime surveillance, seaborne traffic service, to fishery management and military reconnaissance. The increased resolution of available images in the past years makes it more attractive for relevant applications, and the ship detection in high-resolution (HR) remote sensing images has caught increasing attention of researchers.

In recent years, encouraged by the great success of convolutional neural network (CNN) and deep learning based object detection in natural images, many researchers propose to utilize similar methodology for ship detection\cite{zhang2016s, cheng2016learning, long2017accurate, li2017inshore}. However, unlike most of the objects in natural images, the ship targets in remote sensing images are relatively small and less clear. In addition, they have special properties making them harder to be accurately detected than the general objects, and can easily suffer from the degradation by mists, clouds and ocean waves, as well as the impact from the complex background in the harbor. Although compelling results can be achieved by current CNN-based ship detection methods, there is still much room for improvement of the performance, as compared to their counterparts for the object detection in natural images.

One special issue is that the ships in remote sensing images may appear in notably diverse orientations, because of their narrow-rectangle shapes and the high-top viewpoint of imaging from the sky. This is different from the general objects in natural images which are taken from the ground. For those objects, the horizontal bounding boxes are typically used and capable of leading to good detection\cite{Girshick2015Fast, liu2016ssd, redmon2016you, redmon2017yolo9000, Ren2017Faster}. However, for the narrow-rectangle shaped ships in remote sensing images, detecting with horizontal bounding boxes irrespective of their orientations would results in inaccurate detection when they appear with inclined orientations in the image\cite{liu2018arbitrary}, and even worse, it could not separate each individual ship (or would result in miss detection) when they are docked densely beside each other in the harbor\cite{zhang2018toward}. Moreover, the horizontal ground-truth boxes of inclined ships labeled for the training would sometimes contain too much background besides the ships themselves, which may somewhat cause misleading in ship feature learning and extraction.

The more recent researches in the detection algorithm focus on the utilizing of rotated bounding boxes (i.e., the boxes that can be rotated according to the ship orientations) based on CNN and deep learning\cite{liu2017learning, liu2018arbitrary, zhang2018toward}. Rotated bounding boxes are crucial for detecting the ships in diverse orientations. However, most of such algorithms are directly extended from the general ones with horizontal bounding boxes, by introducing an additional variable of orientation into the framework, but the overall strategy remains almost unchanged and all related variables (including the orientation, width, height and center location of the bounding box) are predicted together in one process. The more unknown variables to be considered at the same time would definitely increase the complexity. The current approaches, though effective, are sub-optimal for generating accurate rotated bounding boxes, because of the large uncertainty of the orientation, and its influence on other variables that needs to be handled more properly.

Besides the issue of diverse orientations, other factors or related properties, such as the images not clear enough, background clutters, relatively-small sizes and the shapes with large aspect ratio, make the detection of ships more challenging compared with general objects. Even the well-tuned state-of-the-art object detection methods, for example\cite{Ren2017Faster, redmon2018yolov3}, can not obtain very satisfactory detection results for the ships in remote sensing images. Therefore, more dedicated and innovative treatments are necessary to improve both the detection accuracy and reliability.

In this paper, a novel CNN-based algorithm is proposed to better detect the ships in remote sensing images. For the detection of arbitrary-oriented ships, the existing approaches have to predefine a set of anchors (or called default boxes) in various orientations, on which basis the ship orientation is then be predicted together with other unknown variables via one regression process. In contrast to these approaches, we are able to predict the orientation and other variables independently by creating two regression branches based on the CNN-features with different characteristics. It brings multiple benefits, such as the fewer anchors that must be defined, the easier training, and the increased capacity to achieve more accurate prediction for the orientation as well as other variables. This approach, as will be shown, is able to effectively improve the quality of rotated region proposals, contributing ultimately to generating better detection results with rotated bounding boxes.

Furthermore, we create a feature representation more suitable for the ship detection, which is achieved via a novel feature pooling process on the proposals, named as the multilevel adaptive pooling. The contribution and novelty mainly lie in two aspects. (I) Unlike the typical ROI pooling performed in a fixed pattern, we develop a shape-adaptive pooling which can obtain better spatially-distributed features for the detection of ships with various aspect ratios. (II) We are the first to incorporate multilevel features via the process of pooling, and create a compact feature representation more qualified for the multi-tasks of classification and regression in ship detection. Our approach is different from other detection algorithms using multilevel features\cite{liu2016ssd} or feature pyramid\cite{lin2017feature}. It is built on the motivation that the higher-level features are better at object-level classification, while the lower-level ones are more useful in accurate localization, and we integrate them into one set of representation with a spatially-variant pooling, which enables to meanwhile fully take their advantages.

Our method achieves a considerable boost in the detection performance compared with the state-of-the-art CNN-based methods, without obviously increasing the complexity of network. Experimental results on a variety of images show that it is not only able to generate more accurate rotated bounding boxes, but can also effectively reduce both the false and miss detections. Moreover, we performed a detailed quantitative ablation analysis on the proposed techniques, from which one can get a more comprehensive understanding of why the proposed techniques can work well, and obtain some insights regarding a proper utilizing of multilevel CNN-features (including the feature representation) to improve the detection performance for ship targets in remote sensing images.

The rest of this paper is organized as follows. The related work is introduced in Section II. Section III describes the details of the proposed ship detection method. Experimental results and detailed comparisons are shown in Section IV to verify the superiority of our method. Finally, conclusions are drawn in Section V.

\section{Related Work} \label{section:Related_work}

Ship detection has been an active research topic in the fields of remote sensing for a long period of time. Most of the early ship detection methods rely on geometric elements and some manually-designed features to locate ships from backgrounds. For instance,\cite{lei2007inshore} and\cite{lin2012line} utilize the features of contour and line segment for inshore ship detection. A hierarchical ship detection method was proposed in\cite{zhu2010novel} by using shape and texture features. Generally, the detection with these basic geometric features are susceptible to complex background interference. Some other methods utilize the more prominent features of ship head for preliminary localization. In\cite{li2016novel}, the regions of potential ship heads are first predicted by transforming local pixels into the polar coordinate system, based on which the saliency of directional gradient information is then employed to identify ship body. The ship heads are detected in\cite{liu2013new} by corner features, and then the methods of shape analysis and region growth are used to determine the complete ship region. The method\cite{yang2017ship} first determines the potential ship regions by saliency segmentation, and then the structure-LBP features are used to identify the real ships. Since the manually-designed features can only utilize the low-level information with poor generalization capability, these methods often suffer from the influence of complex background, resulting in either false or miss detection.

Features extracted by the convolutional neural network have stronger semantic information compared with the manually-designed features. They are very useful for the object detection under complex environments. Current CNN-based object detection methods are generally divided into two categories. The first category firstly generate a number of region proposals. On each proposal the classification and regression are then performed to predict the bounding box of object\cite{Girshick2015Fast, Ren2017Faster, lin2017feature}. Faster R-CNN\cite{Ren2017Faster} is one of the most representative methods in this category. It integrates the CNN-feature extraction, proposal generation and the subsequent detection steps into a unified network. The second category deal with object detection as a one-step regression and classification process by discarding the first step of region proposal generation\cite{liu2016ssd, redmon2016you, redmon2017yolo9000, redmon2018yolov3}. For example, SSD (single shot multibox detector)\cite{liu2016ssd} directly predict the bounding boxes together with probability scores  based on the densely distributed default boxes. YOLO (you only look once)\cite{redmon2016you} divides the image into several grids and sets two default boxes in the center location of each grid to perform region classification and bounding box regression.  In general, the one-step methods achieves inferior detection accuracy than the proposal-based methods. Nevertheless, YOLOv3\cite{redmon2018yolov3} (the latest version of YOLO) still achieves impressive performance after several versions of upgradation.

\begin{figure*}[!htp]
\centering
\subfigure{\includegraphics[width=0.93\textwidth]{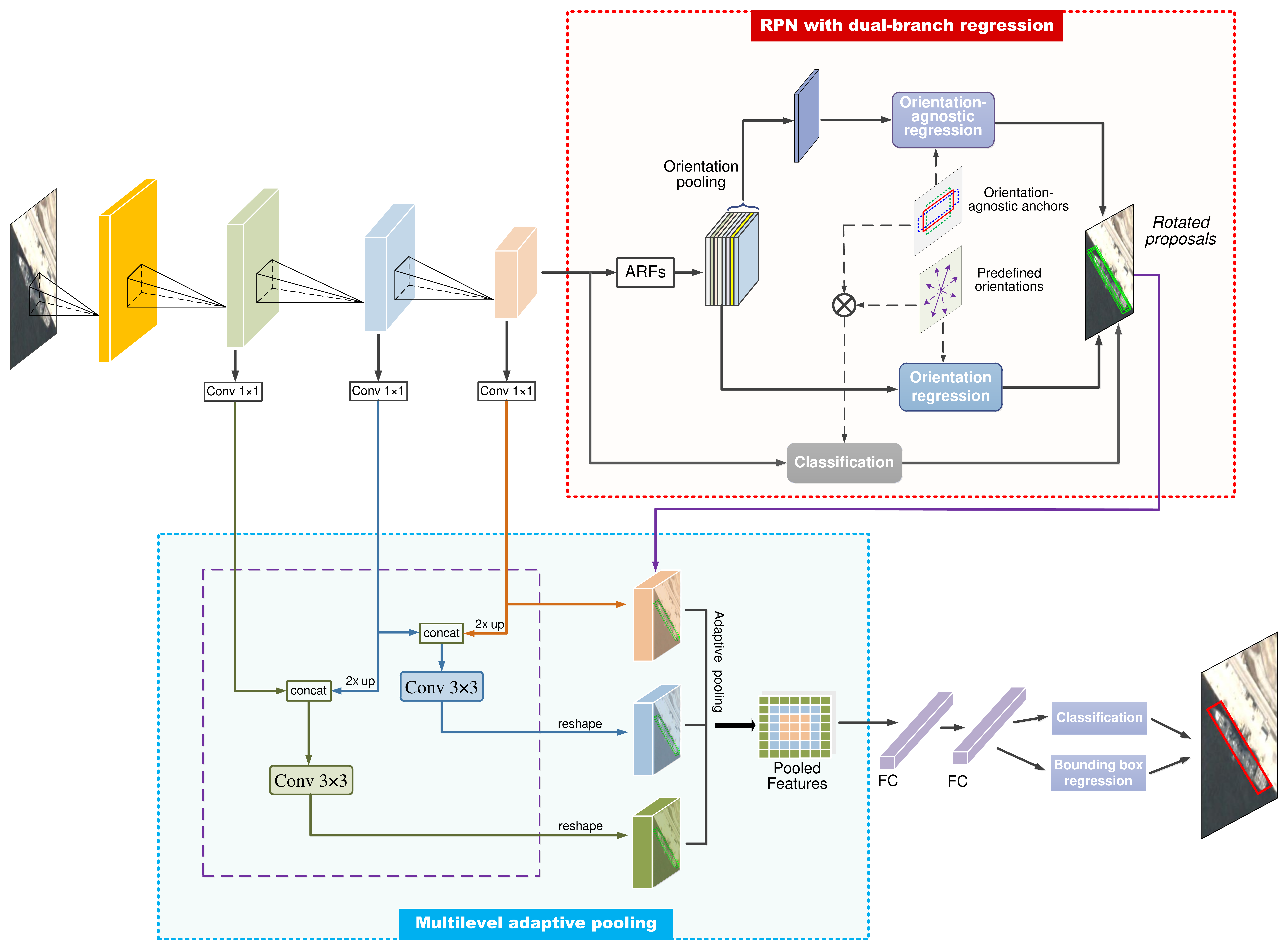}}
\caption{An overview of the proposed method.}\label{Fig:Detection_Pipeline}
\end{figure*}

In the past few years, many CNN-based algorithms are proposed specially to detect the ships or other objects in remote sensing images. For example, a coarse-to-fine method based on CNN was proposed in\cite{li2017object} to better detect diverse objects including the inshore ships in remote sensing images. The SVD algorithm is used to construct a more compact and efficient CNN structure for ship detection\cite{zou2016ship}, in which the candidate regions are extracted through multi-scale features. To improve the localization accuracy, iterative bounding box regression is used in the CNN-based method for ship detection\cite{wu2018inshore}. An end-to-end detection for the ships of various scales was proposed based on region proposal network and the multi-scale feature mapping through hierarchical selection\cite{li2018hsf}. Deng et al.\cite{deng2018multi} presented a more sophisticated detection algorithm for multi-scale objects in remote sensing images, by using multi-scale region proposal coupled with multi-scale feature maps. Such an algorithm, however, comes at a high computational cost. Although the above CNN-based detection methods can typically produce good results, they are unable to accurately locate arbitrary-oriented ships and distinguish the inclined ones that are docked closely beside each other. In order to alleviate this problem,\cite{nie2018inshore} uses a post-processing approach of Soft-NMS\cite{bodla2017soft} to preserve the highly-overlapped boxes in the detection results. In\cite{zhang2019r}, the ROIs are rotated according to their main directions before the detection is performed. However, both the post and preprocessing approaches can not fully resolve the above problem.

Recently, some researchers tried to detect arbitrary-oriented ships in remote sensing images more accurately and reliably with rotated bounding box. In\cite{liu2017learning}, an angle parameter is introduced to define the orientation of the rotatable bounding box, for which the estimation is performed within the typical one-step detection framework. Similarly, to generate the rotated bounding box,\cite{liu2018arbitrary} takes account of the additional angle information in a one-step regression process based on the framework of YOLOv2\cite{redmon2017yolo9000}. One limitation of those methods is that one-step regression is inadequate to accurately estimate the orientation of rotated bounding box, because of its large uncertainty and the more unknown variables that have to be considered in the bounding box regression. To overcome this limitation, many proposal-based detection methods were proposed. For instance,\cite{zhang2018toward} produces the detection results with rotated bounding boxes based on the framework of Faster R-CNN, in which the region proposal network and ROI pooling are generalized to their rotated versions. In\cite{yang2018automatic}, the features generated from a densely-connected version of the feature pyramid\cite{lin2017feature} is used for producing the rotated region proposals and the final bounding boxes. A similar detection approach was proposed in\cite{feng2019towards}, where an alternative feature-pyramid based structure to extract multi-scale context features are employed. To segment the ship targets in remote sensing images,\cite{zhang2019rotationally} introduces a rotated region proposal layer into the framework of Mask R-CNN\cite{he2017mask} and uses an adapted pooling approach modified from ROI-Align\cite{he2017mask} for the feature extraction within the proposals.

Despite the effectiveness of these detection algorithms, the rotated region proposal networks of them are directly extended from the normal horizontal ones by introducing an additional variable of orientation, in which all unknown variables are predicted indiscriminately in one process. This would limit the quality of generated region proposals, and thus impact the final detection results. Another essential issue is the CNN-features that are directly utilized for the ship detection. Although sophisticated approaches, for example, densely-connected feature pyramid\cite{yang2018automatic} or complicated hierarchical connection\cite{feng2019towards}\cite{zhao2019m2det}, can improve the representation power, they are with high complexity and produce the features of high dimensionality, which loses the property of compactness and may increase the risk of over-fitting for the ship targets. In this paper, these problem are well resolved in our method and a considerable boost in the detection performance is rewarded according to the experimental results.

\section{The Proposed Method} \label{section:Proposed_method}
The overall framework of our method is shown in Fig.~\ref{Fig:Detection_Pipeline}. In addition to the typical backbone network, there are two distinctive networks in the proposed algorithm, i.e., dual-branch regression for region proposal and multilevel adaptive pooling. The dual-branch regression, which consists of two independent regression branches (i.e., orientation-agnostic regression and orientation regression), is used to generate high-quality rotated region proposals. Multilevel adaptive pooling is to produce a feature representation more qualified for accurate ship detection in remote sensing images based on the proposals. In the following, we describe them as well as other relevant parts of the proposed algorithm in detail.

\subsection{Rotated Region Proposal Network}\label{subsection:Rotated_RPN}
In many CNN-based detection algorithms, a number of object region proposals are first generated by the region proposal network (RPN). In the absence of any prior information about the object's location and size in the image, the RPN predefines a set of anchors centered on each location of the feature map. Each anchor provides the coarse initial bounding box of object, typically denoted by $(x, y, w, h)$, in which $(x, y)$ represents its center point while $w$ and $h$ represent the width and height, respectively. The refined boxes for object region proposal are then obtained from those anchors by the prediction with the corresponding CNN-features. Recently, it has been discovered that using rotated bounding boxes can effectively improve ship detection performance in remote sensing images\cite{liu2017learning, liu2017rotated}. In this case, the bounding box is expressed as $(x, y, w, h, \theta)$, where the additional variable $\theta$ represents the ship orientation. However, it becomes more complex compared with using typical horizontal bounding boxes, not only due to the increased number of unknown variables, but also the fact that the ship orientation can be arbitrary in images\cite{liu2018arbitrary, zhang2018toward}. To resolve this problem, current methods have to define a set of anchors in various orientations, and predict all of the five variables together in one process. One obvious problem is that it would increase the burden of the shared CNN-features, and consequently limit the potential to acquire more accurate prediction results for individual variables.

\begin{figure}[!t]
\centering
\subfigure{\includegraphics[width=0.39\textwidth]{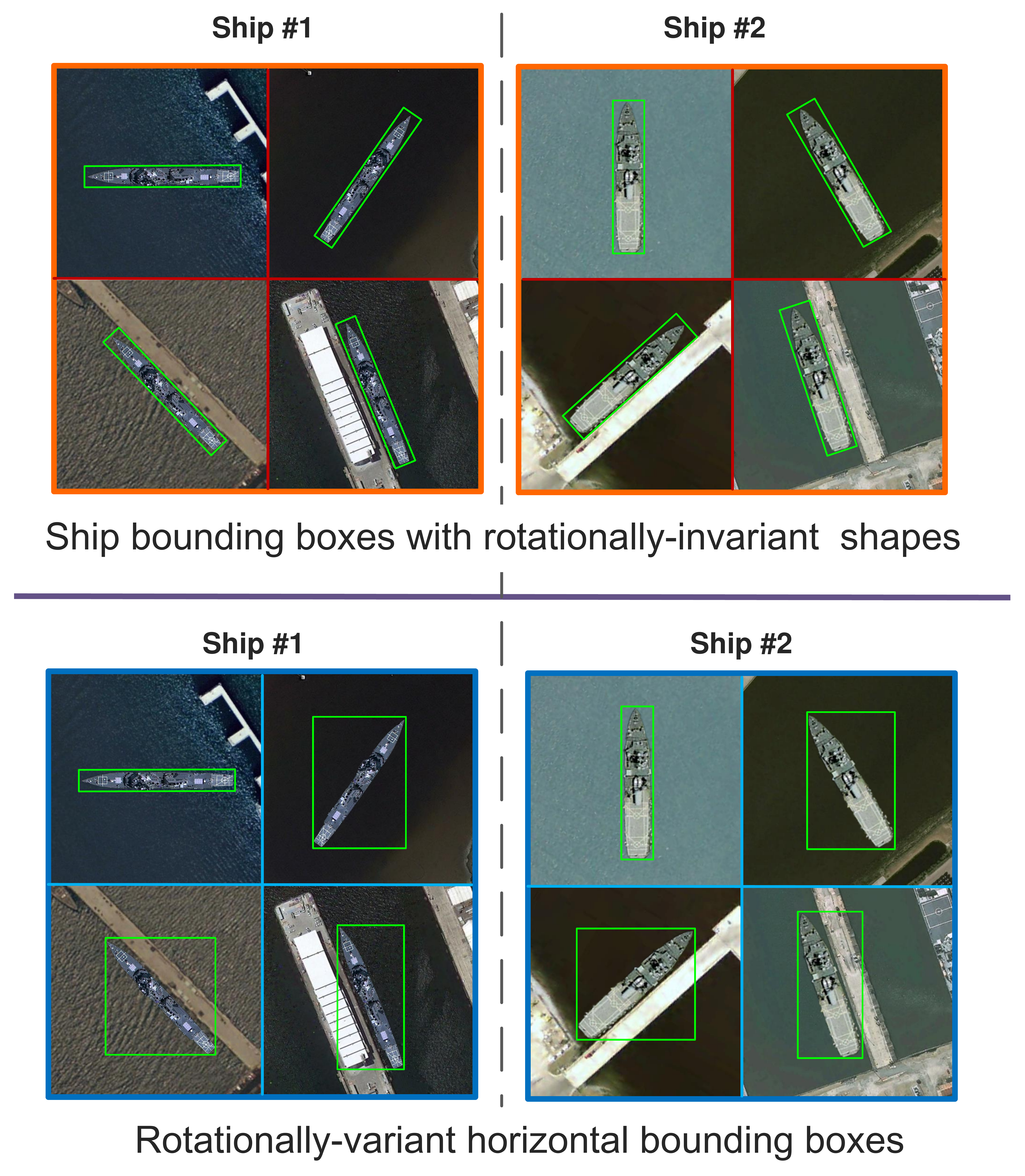}}
\caption{The shape of bounding box is rotationally-invariant for a specific ship in remote sensing images (see the upper illustrations for ship \#1 and \#2, respectively). As a comparison, from the lower part of the figure, we can see that the horizontal bounding boxes of the ships are rotationally-variant (it also indicates that the conventional detection with horizontal bounding boxes can not leverage the useful property of rotation invariance).}\label{Fig:Box_Comparision}
\end{figure}

Fortunately, for a specific ship in remote sensing images, the shape of its bounding box (determined by the variables $(w, h)$) can be assumed to be rotationally invariant (see Fig.~\ref{Fig:Box_Comparision}). This indicates that it is possible to predict $(w, h)$ and $\theta$ independently with separated while more effective approaches. Therefore, unlike existing RPNs that are built on one set of shared CNN-features, we construct two regression branches to predict $(w, h)$ (along with the center point $(x, y)$) and $\theta$ respectively (see Fig.~\ref{Fig:Detection_Pipeline}). Each branch generates particular CNN-features that are more suitable for corresponding regression task.

\textbf{Orientation-agnostic regression: }A rotation-invariant feature module is generated to predict $(x, y, w, h)$ irrespective of the ship orientation $\theta$. The benefit of this module is twofold: first, it helps to decouple the influence of $\theta$ during feature learning, and produce more capable features enabling to better predict $(x, y, w, h)$ independently; second, the ships in any orientation can all be fully used to train the features without the need to differentiate their orientations, and there is also no requirement for the training samples in diverse orientations as compared to current methods\cite{li2018hsf, liu2017rotated, yang2018automatic}.

Specifically, the rotation-invariant feature module consists of multi-oriented response layers and the followed orientation pooling to produce rotation-invariant features. The layers of multi-oriented response are constructed by leveraging Active Rotating Filters (ARFs)\cite{zhou2017oriented}. An ARF is a filter bank ${\{f_0, f_1, f_2 ...  f_{n-1}\}}$ containing an initial convolutional filter $f_0$ and its $2\pi k/N$ clockwise rotated versions ${f_k}(k = 1,2,...,N - 1)$. Given an input $X$, using the ARF we can produce a feature map $F$ consisting of $N$ channels of oriented convolutional response:
\begin{equation}
{F_k} = {f_k} * X,\quad k = 0,1,...,N - 1,
\label{eq:ARF_function}
\end{equation}
where $*$ is the convolution operator, and ${F_k}$ denotes each channel of $F$. Similarly, a set of different feature maps ${F_{\{ i\} }}(i = 1,2,...M)$ can be produced by using the ARFs with differing initial filters. Next, in order to obtain rotation-invariant feature maps, the orientation pooling is performed on each feature map by picking up the maximum oriented response at each location among its $N$ channels.

 Compared with the straightforward way of data augmentation, i.e., rotating the training samples, which often relies on rich convolutional filters to achieve the invariance/tolerance to rotation, using ARFs requires fewer network parameters due to the weight sharing in each ARF, and also decreases both the training cost and the risk of over-fitting. Moreover, the ARF explicitly encodes the orientation-related information into the feature map via the multi-oriented response, which would be beneficial for the orientation regression described below.

\textbf{Orientation regression: }This branch is to focus on the prediction of ship orientation $\theta$. To this end, rotation-sensitive features are used, by sharing the feature maps produced from the ARFs mentioned above. An obvious advantage of such feature maps is that under the transform of rotation, direct changes can be found across the orientation channels, helpful for the subsequent orientation-offset prediction performed by this branch. Another advantage is that they actively comprise the feature response in various orientations, facilitating the efficacy to predict arbitrary orientation of ships through the regression, yet without the necessity of extensive training with a large variety of orientation samples.

Although theoretically it may be possible to predict the orientation through direct regression, we found that the training is hard to converge and it is impossible to achieve the expected performance in the test, because the range of unknown orientation can be as large as ${180^ \circ }$ (two opposite orientations are assumed as the same one). To enable a feasible regression, we resort to predicting the angle offset relative to some predefined orientations which can be much smaller. In our implementation, we define 6 fixed orientations ${\theta _i}(i = 0,1,...,5)$ that are evenly distributed with the interval of ${30^ \circ }$. The regression is then performed to predict each of the corresponding angle offsets ${t_{{\theta _i}}}$, and obtain the final results by correcting ${\theta _i}$ with ${t_{{\theta _i}}}$.

The orientation-agnostic regression branch predicts $(x, y, w, h)$ based on orientation-agnostic anchors (i.e., the horizontal anchors as illustrated in Fig.~\ref{Fig:Detection_Pipeline}). In our implementation we predefine 8 anchors with 4 scales and 2 aspect ratios. This branch predicts the offsets $({t_x}, {t_y}, {t_w}, {t_h})$ relative to each anchor with an 8$\times$4-dimensional output layer. Similarly, the output layer for the branch of orientation regression is 6$\times$1-dimensional. Thus, the number of outputs for the regression is totally 38. As a comparison, current approaches\cite{liu2017learning, liu2018arbitrary, zhang2018toward, liu2017rotated, yang2018automatic} which predict all five variables $(x, y, w, h, \theta)$ together need to generate 2$\times$4$\times$6 anchors that are differing in scale, aspect ratio or orientation. In this case there would be 48$\times$5 outputs for the regression in total.

In addition to the regression branches, a classification branch is used to output the objectness scores for the 6 fixed-orientation clones of each orientation-agnostic anchor, among which only the one with the highest score is selected to be corrected with corresponding offsets $({t_x},{t_y},{t_w},{t_h},{t_{{\theta _i}}})$.

\textbf{Loss function and training: }For training the proposed RPN, the multi-task loss is defined as follows:
\begin{equation}
L({p_c},\textbf{\emph{t}},{t_\theta }) = {L_{\rm{cls}}}({p_c},p_c^ * ) + \lambda p_1^ * ({L_{\rm{reg_1}}}(\textbf{\emph{t}},{\textbf{\emph{t}}^ * }) + p_2^ * {L_{\rm{reg_2}}}({t_\theta },t_\theta ^ * )),
\label{eq:Multi_task_function}
\end{equation}
where ${L_{\rm{cls}}}({p_c},p_c^ * )$, ${L_{\rm{reg_1}}}(\textbf{\emph{t}},{\textbf{\emph{t}}^ * })$ and ${L_{\rm{reg_2}}}({t_\theta },t_\theta ^ *)$ are the classification loss, orientation-agnostic regression loss and orientation regression loss, respectively; $p_1^ *$ and $p_2^ *$ denote the ground-truth labels for the orientation-agnostic anchor and predefined orientation, respectively. The classification and regression terms are balanced by the parameter $\lambda$ ($\lambda  = 1$ is used by default in our method).

\begin{figure}[t]
\centering
\subfigure[]{\includegraphics[width=0.13\textwidth]{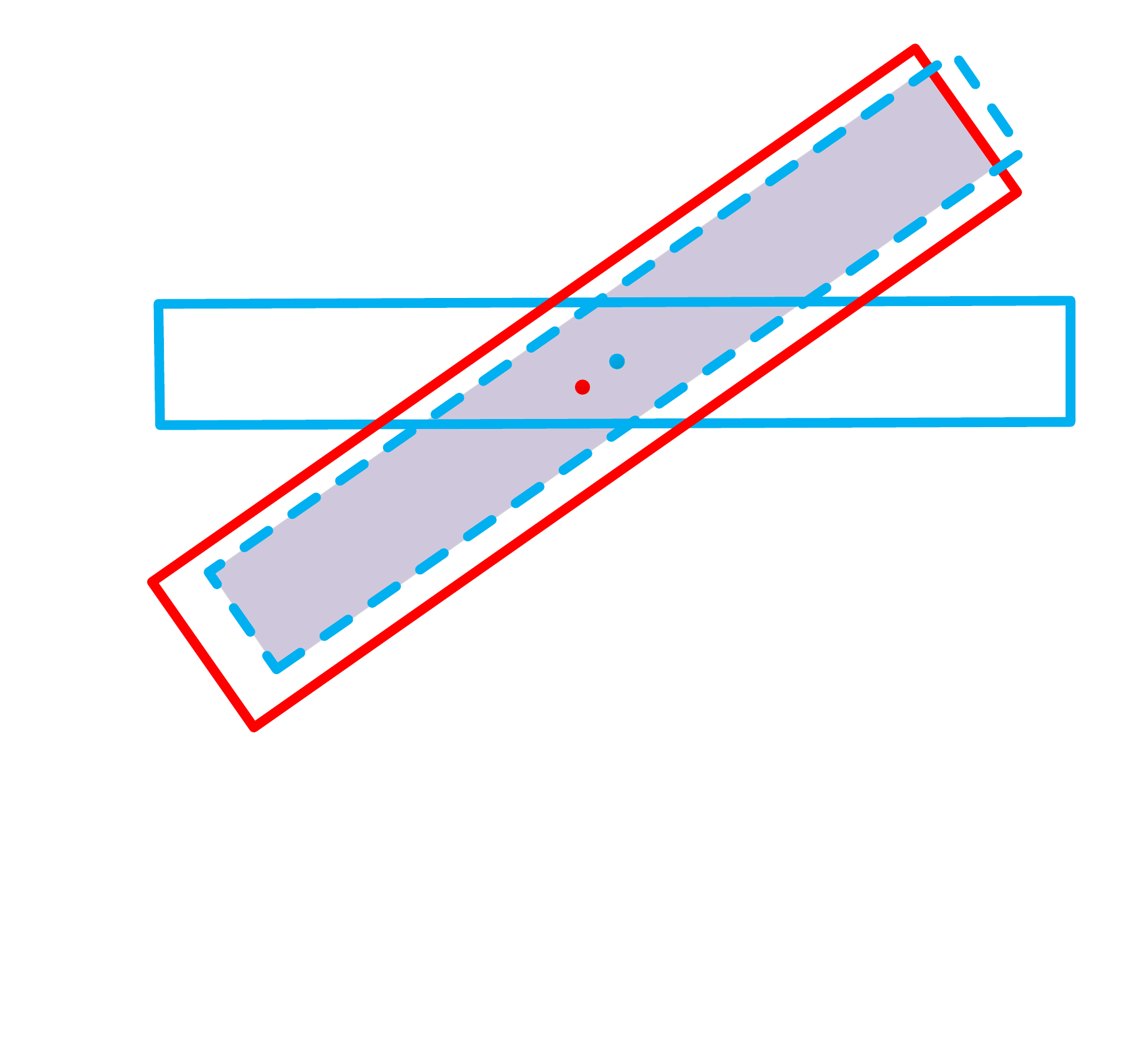}}
\hspace{0.15cm}
\subfigure[]{\includegraphics[width=0.13\textwidth]{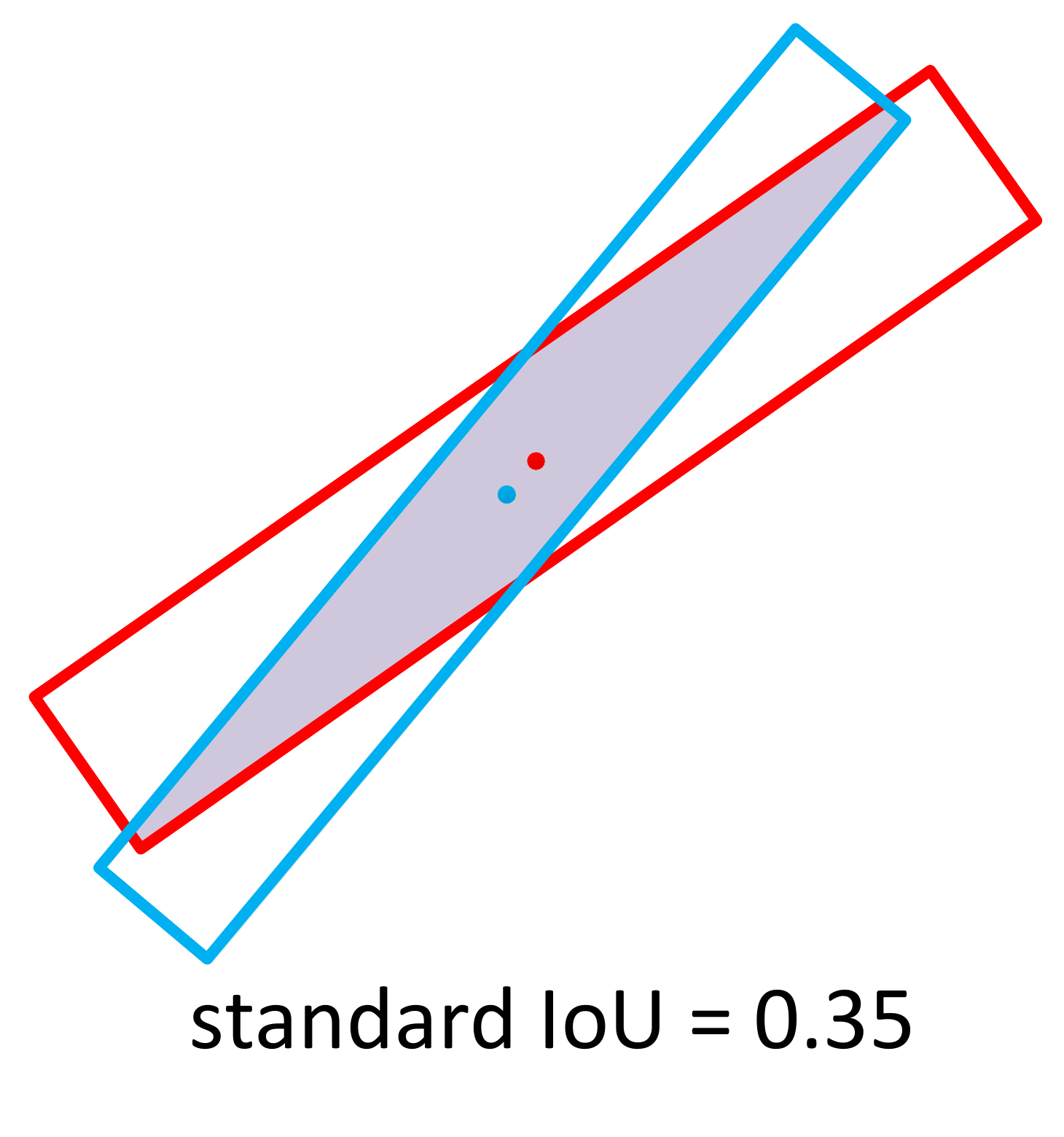}}
\hspace{0.15cm}
\subfigure[]{\includegraphics[width=0.13\textwidth]{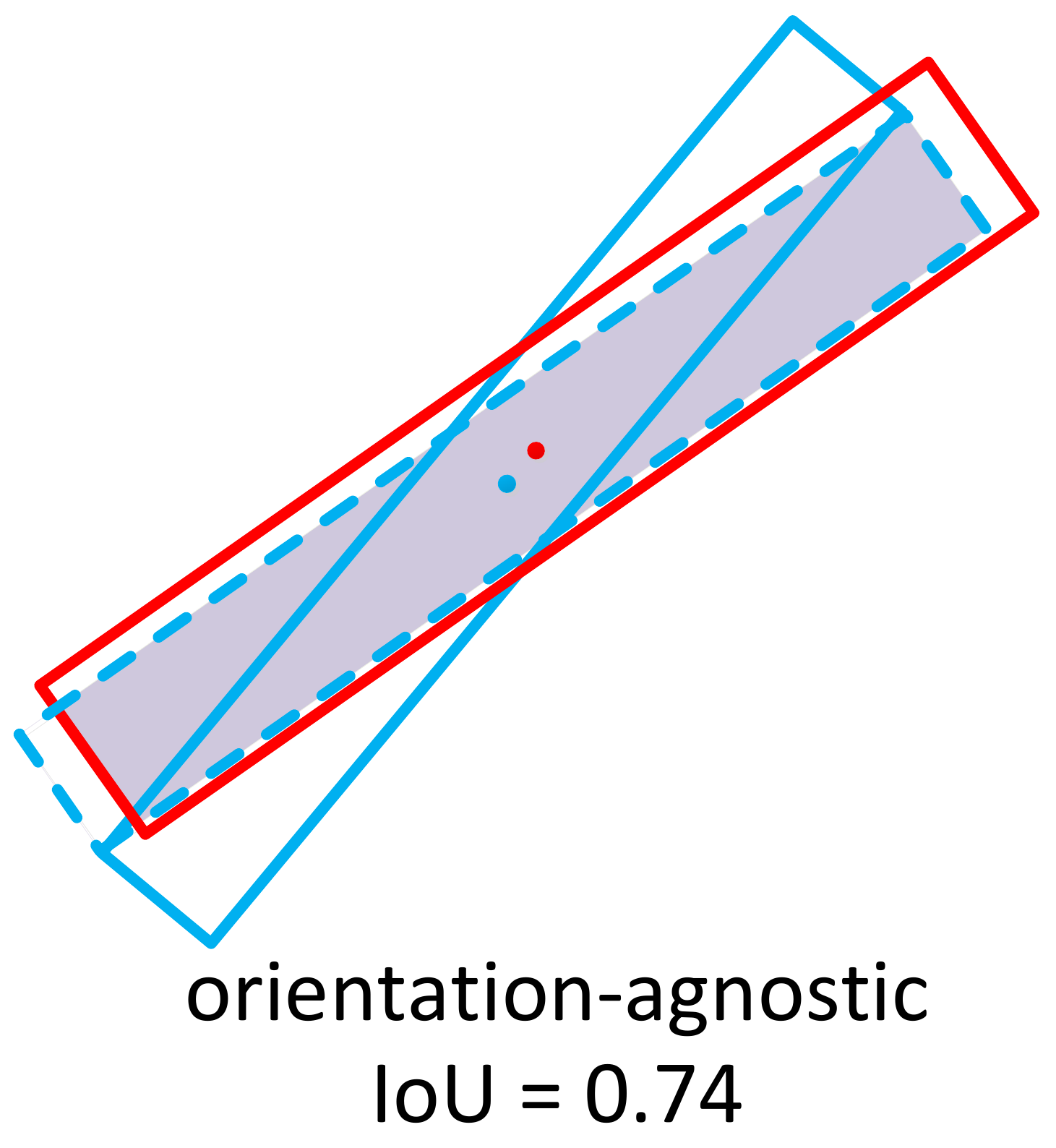}}
\caption{Illustration of the orientation-agnostic IoU and standard IoU, where the shadow area indicates the overlap between the anchor (blue rectangle) and ground-truth box (red rectangle)) that is considered in each case. (a) For an orientation-agnostic anchor and ground-truth box, the IoU is computed by considering the overlap obtained after orientation alignment. (b) The case of standard IoU for an oriented anchor and ground-truth box, in which their direct overlap is considered. (c) The case of orientation-agnostic IoU for the oriented anchor and ground-truth box.}\label{Fig:IOU}
\end{figure}

We assign the label $p_1^ *$  for orientation-agnostic anchors based on their Intersection-over-Union (IoU) overlap with the ground-truth boxes like typical approaches\cite{liu2017learning, zhang2018toward, yang2018automatic}. However, in our case we need to define an alternative IoU, named as orientation-agnostic IoU, to achieve this goal. The orientation-agnostic IoU is computed by taking no account of the orientation difference between the anchor and ground-truth box, i.e., they should be aligned in orientation when the IoU is computed (see Fig.~\ref{Fig:IOU}(a)). In our method, two kinds of orientation-agnostic anchors are assigned with the positive label $p_1^ *  = 1$: (I) the orientation-agnostic anchor/anchors having the highest orientation-agnostic IoU overlap with a ground-truth box, or (II) an orientation-agnostic anchor having an orientation-agnostic IoU overlap higher than 0.7 with any ground-truth box. The others are assigned with the negative label $p_1^ *  = 0$, so that their corresponding regression loss can be disabled in Eq.~(\ref{eq:Multi_task_function}). For a similar purpose, we assign the positive label $p_2^ *  = 1$ to a predefined orientation if its deviation from the orientation of any ground-truth box at the location is smaller than $\pi /6$, and assign the negative label $p_2^ *  = 0$ otherwise.

The regression outputs the offsets $\textbf{\emph{t}}=({t_x},{t_y},{t_w},{t_h})$ and $t_\theta$ from the two independent branches, respectively, and we have:
%They are formulated as follows:
\begin{equation}
\begin{array}{l}
{t_x} = (x - {x^a})/{w^a},{t_y} = (y - {y^a})/{h^a}\\
{t_w} = \log (w/{w^a}),{t_h} = \log (h/{h^a})\\
{t_\theta } = \tan (\theta - {\theta ^a}),
\end{array}
\label{eq:Output_t}
\end{equation}
in which $x$, $x^a$ are for the orientation-agnostic anchor and the predicted box, respectively (likewise for $y,w,h$), and $\theta$, $\theta ^a$ denote the predefined and predicted orientations, respectively. Given the ground-truth offsets $\textbf{\emph{t}}^*=({t_x^*},{t_y^*},{t_w^*},{t_h^*})$ and $t_\theta ^*$, we employ the smooth-$L_1$ loss\cite{Girshick2015Fast} for the regression:
\begin{equation}
{L_{\rm{reg_1}}}(\textbf{\emph{t}},{\textbf{\emph{t}}^ * }) = \sum\nolimits_{i \in \{ x,y,w,h\} } {{\rm smooth}{_{{L_1}}}(t_i^* - {t_i})},
\label{eq:Reg_1}
\end{equation}
\begin{equation}
{L_{\rm{reg_2}}}({t_\theta },t_\theta ^*) = {{\rm smooth}{_{{L_1}}}(t_\theta^* - {t_\theta})},
\label{eq:Reg_2}
\end{equation}
where
\begin{equation}
\begin{array}{l}
{t_x^*} = (x^* - {x^a})/{w^a},{t_y^*} = (y^* - {y^a})/{h^a}\\
{t_w^*} = \log (w^*/{w^a}),{t_h^*} = \log (h^*/{h^a})\\
{t_\theta ^*} = \tan (\theta ^* - {\theta ^a}),
\end{array}
\label{eq:Real_t}
\end{equation}
in which, $x^*$, $y^*$, $w^*$, $h^*$ and $\theta^*$ denote the corresponding ground-truth values.

${L_{\rm{cls}}}({p_c},p_c^ * )$ calculates the classification loss for the oriented anchor obtained from the combination of the given orientation-agnostic anchor $(x,y,w,h)$ and predefined orientation $\theta$. ${L_{\rm{cls}}}({p_c},p_c^ * )$ is defined as:
\begin{equation}
 {L_{\rm{cls}}}({p_c},p_c^ * ) =  - {(1 - \hat{p}_c )^\gamma }\log (\hat{p}_c ),
\label{eq:Lcls}
\end{equation}
where
\begin{equation}
\hat{p}_c = \left\{ \begin{array}{l}
{p_c}{\rm{        }} \qquad \ \  {\rm if}\ p_c^* = 1\\
1 - {p_c}{\rm{   }} \quad  {\rm otherwise},
\end{array} \right.
\label{eq:pc}
\end{equation}
in which $p_c^*$ denotes the ground-truth label for the oriented anchor ($p_c^* = 1$ if it is positive, and $p_c^* = 0$ otherwise), and $p_c$ is the predicted probability for it. Different from the widely used cross entropy loss, there is an additional modulating factor $(1 - \hat{p}_c)^\gamma$\cite{lin2017focal} in Eq.~(\ref{eq:Lcls}). The modulating factor $(1 - \hat{p}_c)^\gamma$ down-weights the loss of well-classified examples with large $p_c^*$, and thus focus the loss on hard samples that have been misclassified ($ \gamma =2$ is used in our implementation).

To assign the ground-truth label for the oriented anchor, typical approaches\cite{liu2017learning, zhang2018toward, yang2018automatic} use the standard IoU, which is calculated based on the direct overlap between the oriented anchor and ground-truth box (see Fig.~\ref{Fig:IOU}(b)). However, the standard IoU is sensitive to the orientation deviation between the oriented anchor and ground-truth box because of their narrow-rectangle shapes. As shown in Fig.~\ref{Fig:IOU}(b), when the orientation deviation ${\Delta _\theta }$ is only ${15^ \circ }$, the standard IoU can drop to 0.35, whereas the orientation-agnostic IoU (${\Delta _\theta=0}$) is 0.74 (see Fig.~\ref{Fig:IOU}(c)). Because of this, the standard IoU is not so effective as expected in determining the ground-truth label for the oriented anchor of ship targets. Therefore, instead of using the standard IoU, we assign the ground-truth label $p_c^*$ by taking into consideration the orientation-agnostic IoU (denoted by $p_{iou}$) and orientation deviation separately. In our method, the relevant criterions to identify the positive oriented anchor are $p_{iou}>0.7$ (or the one having the highest $p_{iou}$ with a ground-truth box) and ${\Delta _\theta }<\pi /6$. This is equivalent to use $p_c^* \leftarrow p_1^* p_2^*$ to determine the positive and non-positive oriented anchors. Among the non-positive oriented anchors, only the definitely-negative ones with $p_{iou}<0.3$ or ${\Delta _\theta } > \pi /3$ contribute to the related training objective.

\subsection{Multilevel Adaptive Pooling} \label{subsection:Pooling}
\begin{figure}[t]
\centering
\subfigure[Typical ROI pooling]{\includegraphics[width=0.37\textwidth]{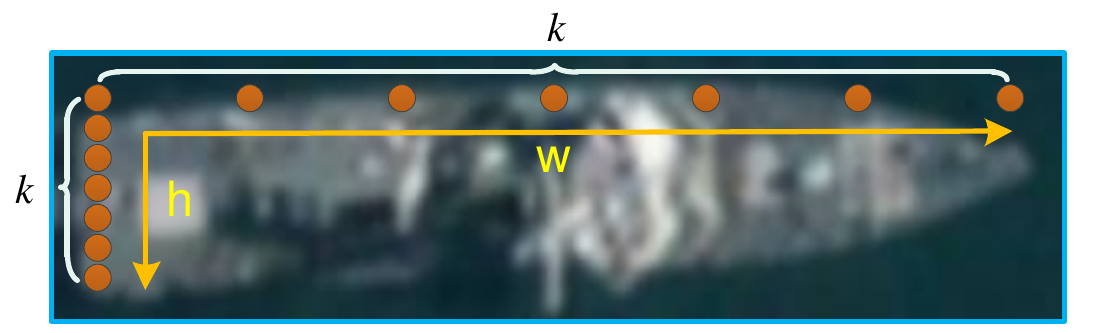}}\\
\subfigure[Shape-adaptive pooling]{\includegraphics[width=0.37\textwidth]{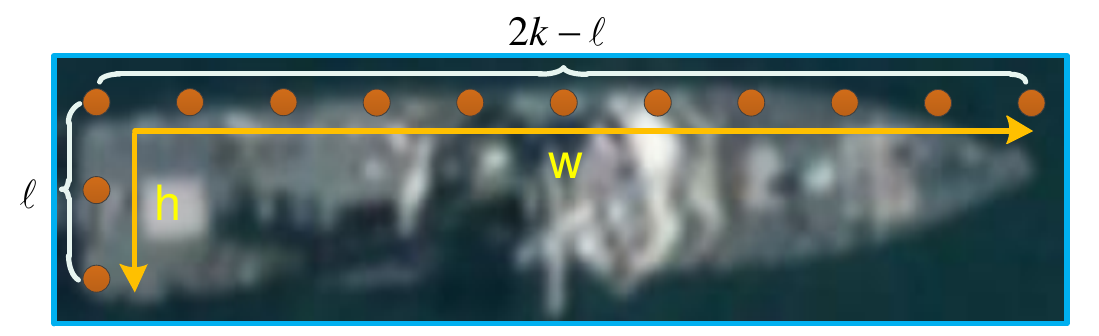}}
\caption{Comparison of the typical ROI pooling and the proposed shape-adaptive pooling. Note that in (b) $\ell$ varies according to the shape (i.e., aspect ratio) of the box.}\label{Fig:Adaptive_Pooling}
\end{figure}

In the second stage, further refinement of proposals is achieved based on the features from each proposal. For this purpose, the first step is to pool the features in different proposals to a fixed size, usually termed as ROI (region of interesting) pooling\cite{Ren2017Faster, zhang2018toward, liu2017rotated, ma2018arbitrary}. However, to obtain a fixed size of $k \times k$ features, the typical ROI pooling uniformly outputs the same number of $k$ feature samples along the width and height directions regardless of the aspect ratio of proposal. For the ship region proposals which are mostly in narrow-rectangle shapes (i.e., with large aspect ratio $w/h$), it would result in severely uneven pooling of features along the two directions, i.e., one is extremely dense while the other is very sparse (Fig.~\ref{Fig:Adaptive_Pooling}(a)). The resultant feature distribution emphasized on the narrow side would be apparently unfavourable to accurate ship detection. The ROI pooling designed for general object detection, however, has not taken account of this problem. In this paper, we propose an adaptive pooling to get the feature representation that is better distributed according to the shape of proposal (Fig.~\ref{Fig:Adaptive_Pooling}(b)). This approach would particularly function well in our method, since the proposed RPN is able to provide high-quality proposals for the ship targets.

\begin{figure*}[t]
\centering
\subfigure[]{\includegraphics[width=0.55\textwidth]{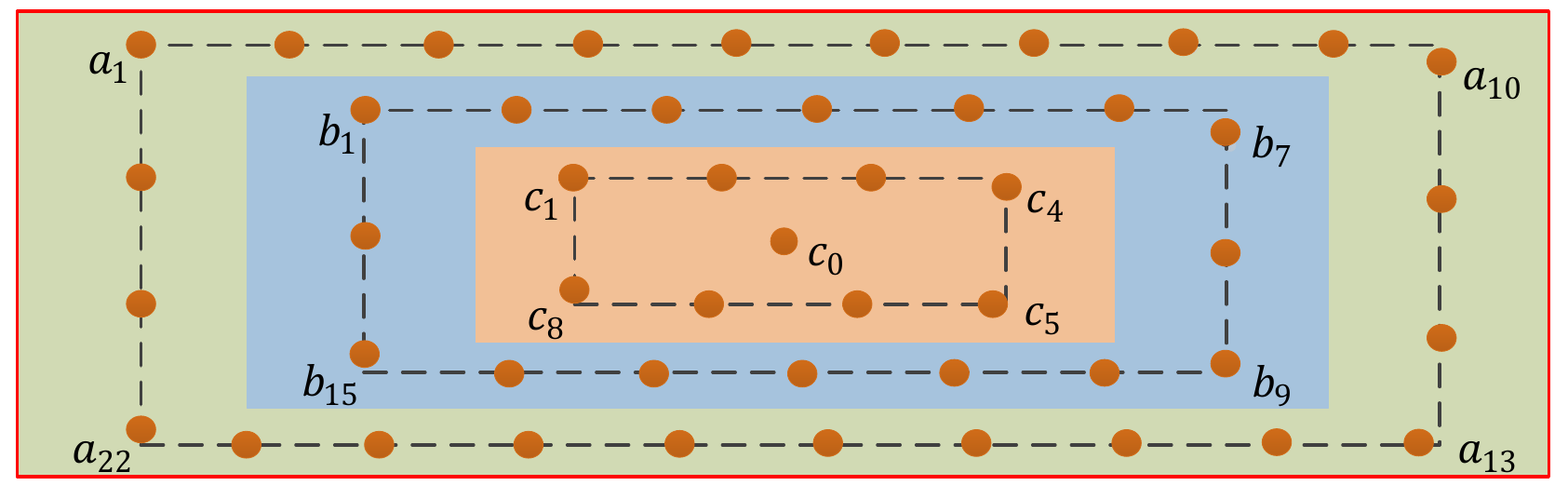}}
\hspace{1.5cm}
\subfigure[]{\includegraphics[width=0.25\textwidth]{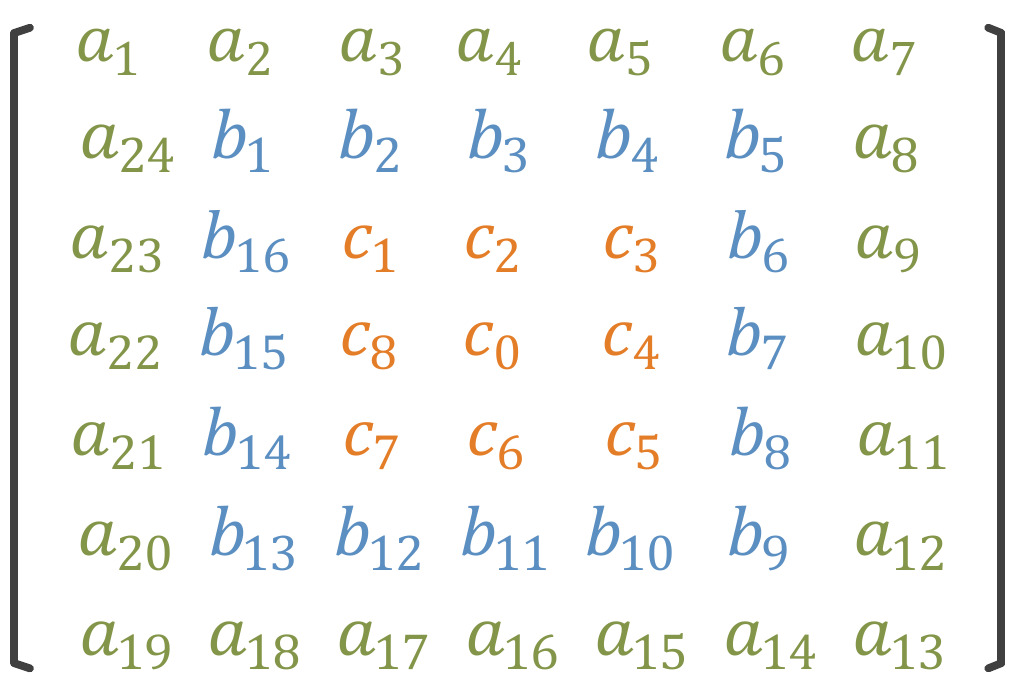}}
\caption{Illustration of the multilevel shape-adaptive feature pooling on the proposal (in this example the size of output features is 7$\times$7). (a) The feature samples on different circles are shape-adaptively pooled from the feature maps of different levels. (b) The output features with fixed size.}\label{Fig:Sampling}
\end{figure*}

Furthermore, we propose to incorporate multilevel features through the pooling, aiming at generating the features more qualified for the subsequent multi-tasks of classification and regression. Generally, high-level convolutional features can be more insensitive to intra-class variations, including geometric distortion and small shift, which is beneficial for dealing with the classification task. To a certain extent, however, this contradict the demand of regression in object detection, whose task is to locate the object's bounding box as accurately as possible. On the contrary, low-level features encoded in the shallower layers of convolutional network, such as the features of edges, texture and corners, are more spatially-sensitive. They are better at localization but less semantically meaningful for object-level classification. Although previous work\cite{li2018hsf, redmon2018yolov3, yang2018automatic} also found the importance of using multilevel features for object detection, the underlying motivation as well as the purpose in this paper is quite different. To the best of our knowledge, we are the first to advocate the incorporation of multilevel features for promoting simultaneous object localization and classification, and achieve this goal via a novel pooling process.

More specially, unlike the conventional pooling that is conducted in rows and columns within the proposal, we perform a ring-like pooling as shown in Fig.~\ref{Fig:Sampling}. It makes the pooled features spatially arranged along different square-shaped circles (denoted by the enclosed dashed-lines in Fig.~\ref{Fig:Sampling}). The basic idea is to let the feature samples along outer circles pooled from lower-level feature maps, while the inner feature samples come from the feature maps of higher-level. Such spatially-variant multilevel pooling helps to create a compact representation, which enables to meanwhile take full advantage of both lower and higher-level features. This is based on the consideration that, within the region of an object, the positions farther away from the center are typically more sensitive to the object's geometric transformation (e.g., in scale, aspect ratio and orientation). At those positions the lower-level features are preferred, helping to maximum their advantage in object localization. For relatively indiscernible ships in remote sensing images, in particular, the only prominent cues served for accurate localization of them are usually their low-level features of boundary, which are expected to be incorporated by the outermost circle of pooling. Conversely, the higher-level features, which have larger receptive fields and focus more on object-level classification, should be pooled from the inner areas.

Along each circle, the aforementioned shape-adaptive pooling is performed. Specially, instead of dividing the neighborhood into bins and aggregating the feature values of each bin using max or average operation as usual, we only sample the value of feature at the center of each bin by bilinear interpolation. It consequently leads to a set of sampled points evenly spaced along each circle as shown in Fig.~\ref{Fig:Sampling}(a) (denoted by the brown dots). The underlying consideration is that interpolating a single value of the center is nearly as effective as the max or average pooling\cite{he2017mask}, while the calculation is sensibly simplified, and can (like the ROI-Align\cite{he2017mask}) avoid the harsh quantization which introduces harmful misalignment between the proposal and extracted features. This process is made sure to totally output the feature values of fixed size, e.g. $7 \times 7$ in our implementation (see Fig.~\ref{Fig:Sampling}(b)), among which the three groups of values $a_1-a_{24}$, $b_1-b_{16}$ and $c_0-c_8$ are sampled from different circles, respectively. The space between neighboring sampled points along each circle depends on the circle's perimeter, and the distribution of each group of the sampled points varies according to its aspect ratio. It is such sampling in a circle that facilitates the performance of shape adaptation. To ensure an even distribution of the three circles within the proposal, in our implementation their widths \& heights are set to be $\frac{6}{7}$, $\frac{4}{7}$ and $\frac{2}{7}$ of those of the proposal, respectively.

The above pooling shares partly a similar goal with the deformable ROI pooling\cite{dai2017deformable}, i.e., augmenting the distribution of sampling locations within the proposal. Our approach is however based on a readily assigned sampling pattern without the learning and computation of the offsets (which may vary significantly for different instances), and able to properly accommodate multilevel features.

The input feature maps for different circles should also be different as aforementioned. However, we found that directly input different level of raw feature maps from deep ConvNet would suffer degradation in performance, which is probably because of the large semantic gaps between different levels. To address this problem, as shown in Fig.~\ref{Fig:Detection_Pipeline}, we `smooth' the gaps through a $3 \times 3$ convolution between the higher and lower levels (prior to which the 2x upsampling of higher level needs to be performed).

\subsection{Implementation Details} \label{subsection:Impletation_details}
We use all convolutional layers in VGG16-Net\cite{simonyan2014very} as the backbone network, which followed by the proposed RPN and multilevel adaptive pooling. There are 3 successive multi-oriented response layers in our RPN, each with 40 Active Rotating Filters for 8 orientations. The four scales of the orientation-agnostic anchor are 32, 64, 128 and 256, and the two aspect ratios are 1:4 and 1:7. Up to 2000 and 300 rotated region proposals generated by the proposed RPN are reserved for training and test, respectively. Similar to some other methods\cite{ma2018arbitrary, ma2019ship}, we enlarge both the width and height of the generated proposals by a factor of 1.2 to utilize more contextual information. Next, multilevel adaptive pooling outputs feature maps of $7 \times 7 \times 256$ for each rotated region proposal. To facilitate the pooling operation, multilevel features are unified to a same size after being smoothed. In our implementation, we achieve this by the reshape operation which converts larger feature maps to a set of smaller ones. To obtain the final refined bounding box, after the pooling, we set two 1024-d fully-connected (FC) layers to perform the classification and regression like other proposal-based ship detection methods\cite{zhang2018toward, yang2018automatic}. In the end, NMS post-processing with a threshold of 0.2 is used to remove the redundancy.

\section{Experiments}\label{section:Experiments}

\subsection{Dataset and Training Details}
We train and test the proposed method on a ship dataset with 1300 images which contains 5386 ships in various background collected from Google Earth. For a fair comparison with the object detection methods using horizontal bounding boxes, all the ships are annotated by both horizontal bounding boxes and rotated bounding boxes. The entire dataset is randomly divided into three parts (training set, validation set and test set) with a ratio of 6:1:3. Data augmentation is performed by flipping each image in horizontal and vertical reflections to triple the images for training.

Besides the ImageNet pre-trained backbone network, all the other learnable layers are initialized from a zero-mean Gaussian distribution with standard deviation 0.01. The input image is resized such that its shorter side has 600 pixels. The network is trained with Adam optimizer on GTX1080ti GPU with 2 images per mini-batch and a total of 256 anchors with a foreground-to-background ratio of 1:3 per image. During training, we use a learning rate of 0.001 for 40k mini-batches, and 0.0001 for the next 40k mini-batches. We also use a momentum of 0.9 and a weight decay of 0.0005. Moreover, we take the IoU threshold of 0.6 for proposal classification. That is, the proposals having IoU overlap with any ground truth box larger than 0.6 are regarded as positive samples, and the others are negative samples. Compared with the typical value of 0.5, a higher threshold can help to preserve more accurate proposals for the detection.

\subsection{Experimental Analysis}
We mainly use the average precision (AP) and precision-recall curve to evaluate the performance of different methods. The AP is the average value of precisions based on different recalls. The precision and recall indicators are formulated as follows:
\begin{equation}
{\rm{Precision = }}\frac{{TP}}{{TP + FP}},
\label{eq:Precision}
\end{equation}
\begin{equation}
{\rm{Recall = }}\frac{{TP}}{{TP + FN}}.
\label{eq:Recall}
\end{equation}
Here, $TP$, $FP$, and $FN$ denote the number of true-positives, false-positives and false-negatives respectively. A true-positive means that the IoU overlap between the predicted bounding box and the ground truth box is higher than 0.5.

In addition, we use the average recall (AR)\cite{hosang2015makes} to quantitatively evaluate the quality of region proposals, which calculates the average recall for fixed number of proposals between IoU 0.5 to 1.

\emph{1) Evaluation of the proposed RPN}: To verify the superiority of the proposed RPN with dual-branch regression, we conduct the comparison with a baseline model in which all variables of rotated proposals are predicted together in a single regression process. It is a direct extension of the typical horizontal RPN by simply introducing the unknown variable of $\theta$ into the original formalization, and widely used in the existing ship detection algorithms with rotated bonding box\cite{liu2017learning, liu2018arbitrary, zhang2018toward, liu2017rotated, yang2018automatic}. For a fair comparison, the predefined scales, aspect ratios and orientations in the baseline are all consistent with those in the proposed RPN.

We conduct several experiments to evaluate the quality of proposals generated by these two models, and the experimental results are shown in Table~\ref{table:RPN}. We report the ARs for 100, 500 and 1000 top-scoring proposals per image (denoted by $\rm{AR^{100}}$, $\rm{AR^{500}}$ and $\rm{AR^{1k}}$, respectively). The higher scores of the proposed RPN on all the ARs indicate that the dual-branch regression is able to effectively improve the performance of region proposal for ships. In addition, from Table~\ref{table:RPN} it can also be seen that the mean IoU (only positive proposals having an IoU overlap higher than 0.5 with any ground-truth box are involved here) is obviously improved by the proposed RPN compared with the baseline model.

\renewcommand\arraystretch{1.5}
\begin{table}[!htp]
\centering
\caption[]{Performance evaluation of the proposed RPN}\label{table:RPN}
\setlength{\tabcolsep}{3mm}{
\begin{tabular}{lcccc}
\hline
\hline
      Model    &  ${\rm{AR}}^{100}$  &  ${\rm{AR}}^{500}$  &  ${\rm{AR}}^{1k}$   &  IoU   \\
\hline
  baseline    &  76.2\%     &   81.5\%    &   85.3\%    &   0.671     \\

  proposed RPN     &  79.6\%     &   84.2\%    &   87.5\%    &   0.734     \\
\hline
\hline
\end{tabular}}
\end{table}

\emph{2) Evaluation of multilevel adaptive pooling}: To evaluate the effectiveness of multilevel adaptive pooling, we constructed the detection methods using different types of pooling operations for comparison. The other parts of these compared detection methods are all kept consistent except the subnetwork of pooling (the proposed RPN is used in all of them for region proposal). The comparison results are provided in Table~\ref{table:Pooling}, where we use the index of AP to evaluate the performance of detection based on different types of pooling operations. In the comparison, the benchmark (the first row) is the typical ROI pooling on the single-level features (i.e., the highest-level features from the backbone network). This type of pooling operates on regular $k \times k$ rotated bins, which can be seen as a rotated version of the original ROI pooling \cite{Ren2017Faster} and is used in many current detection algorithms\cite{zhang2018toward, liu2017rotated, ma2018arbitrary}. We can see that using the proposed multilevel adaptive pooling (the fourth row) is able to obtain obviously higher value of AP in the detection.

For more detailed evaluation, we also compared other different types of pooling in Table~\ref{table:Pooling}, i.e., the adaptive pooling on single-level features (the second row) and the typical pooling on multilevel features (the third row). We can see that they produce comparable improvement over the benchmark. It is interesting to notice that the adaptive pooling brings significantly higher improvement on multilevel features (row \#4 vs. row \#3) than that on single-level features (row \#2 vs. row \#1). This can probably be explained as following. The single level only contains the highest-level features, which are relatively insensitive to spatial distortion. This characteristic alleviates the problem of uneven pooling in the spatial domain caused by the typical method. However, when one wants to take advantage of multilevel features, since more lower-level features are involved, the adaptive pooling which can generate better spatially-distributed features would show more superiority. In this case, the multilevel features can also better release its potential for the detection (row \#4 vs. row \#2).

\begin{table}[!htp]
\centering
\caption[]{Quantitative evaluation of multilevel adaptive pooling}\label{table:Pooling}
\begin{tabular}{ccc}
\hline
\hline
      Pooling method    &  Features of pooling  &  AP   \\
\hline
  typical    &   single-level    &   87.9\%     \\

  adaptive    &   single-level    &   88.3\%     \\

  typical    &   multi-level    &   88.4\%     \\

  adaptive    &   multi-level    &   \textbf{89.2\%}     \\

\hline

  adaptive    &   multi-level (channel concat)    &   88.6\%     \\

  adaptive    &   multi-level (reverse connect)    &   88.4\%     \\
\hline
\hline
\end{tabular}
\end{table}

\emph{3) Different arrangements of multilevel features}: In this paper, we perform a ring-like spatially-variant pooling, in which the outer feature samples are pooled from the lower-levels, while the inner feature samples come from the higher-levels. To verify the superiority of this arrangement for multilevel features, we compare it with the other two possible arrangements in Table~\ref{table:Pooling}. The first one is the direct concatenation of multilevel features in channel-wise (denoted by `channel concat' in Table~\ref{table:Pooling}) after the pooling. In this case, the lower and higher-level pooled features are fully overlapped at each location, and the total number of feature channels are increased. The second one is the reverse arrangement opposite to our approach, in which the inner feature samples come from the lower-levels while the outer ones are pooled from the higher-levels. From Table~\ref{table:Pooling}, it can be seen that the above two arrangements both cause worse average detecting precision. Though not very apparent, we can see that the reverse one, in particular, is inferior to the one with channel concatenation.

\begin{figure}[!htp]
\centering
\subfigure{\includegraphics[width=0.40\textwidth]{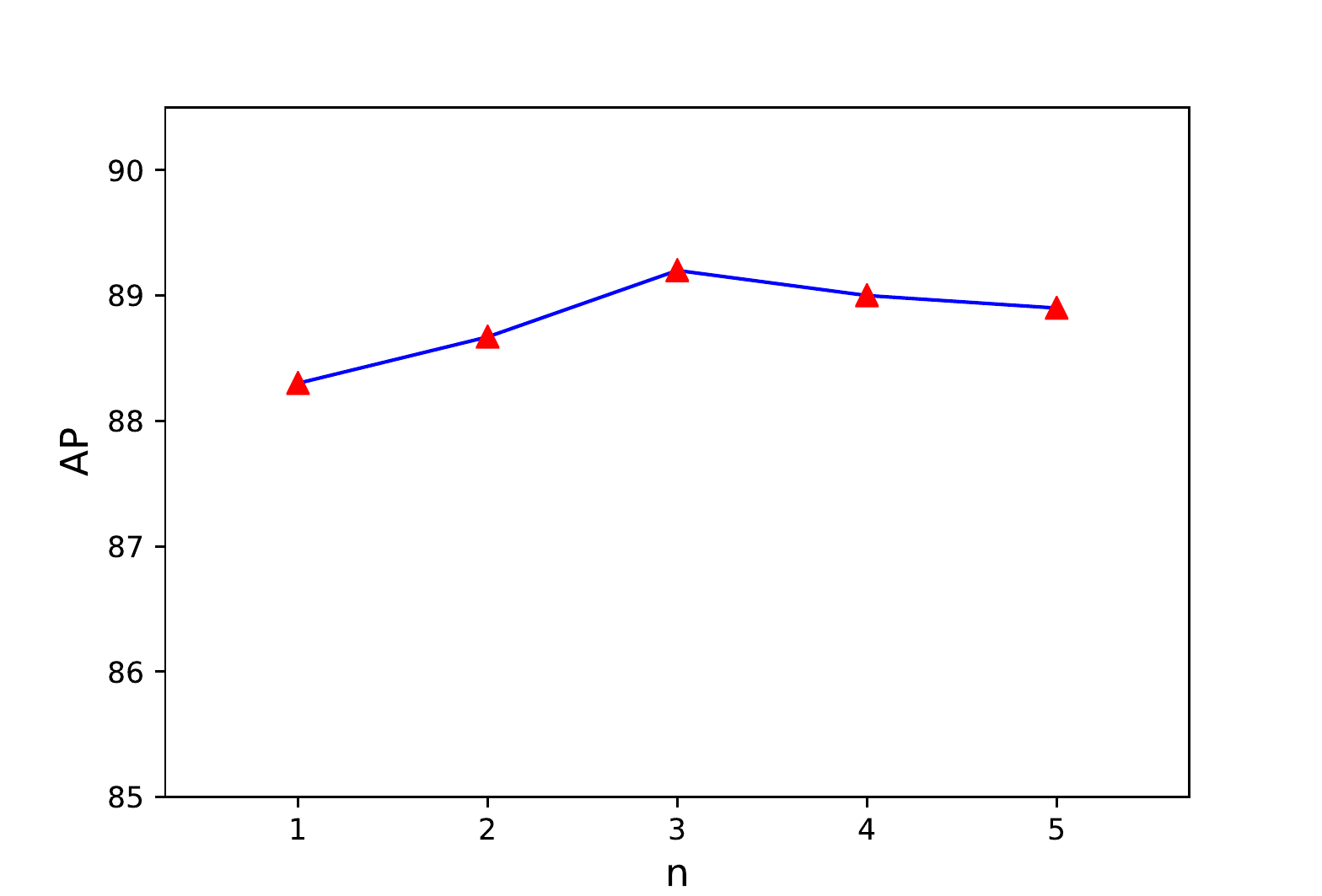}}
\caption{AP versus number of feature levels. }\label{Fig:feature_level}
\end{figure}

\emph{4) Different numbers of feature levels}: The backbone network produces a total of five different level features according to the size of feature maps. To assess the influence of the number of used feature levels, the plot of AP versus number of feature levels is provided in Fig.~\ref{Fig:feature_level}, where the number of used feature levels $n$ ranges from 1 to 5. In each case, only the top $n$ levels of the backbone features are employed. From the plot in Fig.~\ref{Fig:feature_level}, we can see that more than one level would definitely improve the detection performance. However, when sufficient levels are used (in our method, the right number is 3), involving more lower-level features would cause a degradation in the performance. This reveals that the extremely lower-level features are not only useless, but might cause misleading in the learning-based ship detection.

\begin{figure*}[!htp]
\centering
\subfigure{\includegraphics[width=0.19\textwidth]{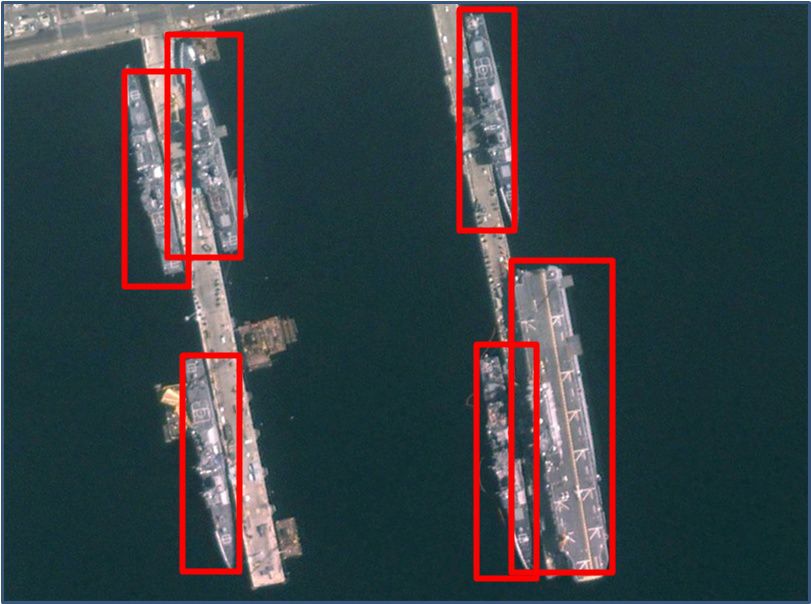}}
\subfigure{\includegraphics[width=0.19\textwidth]{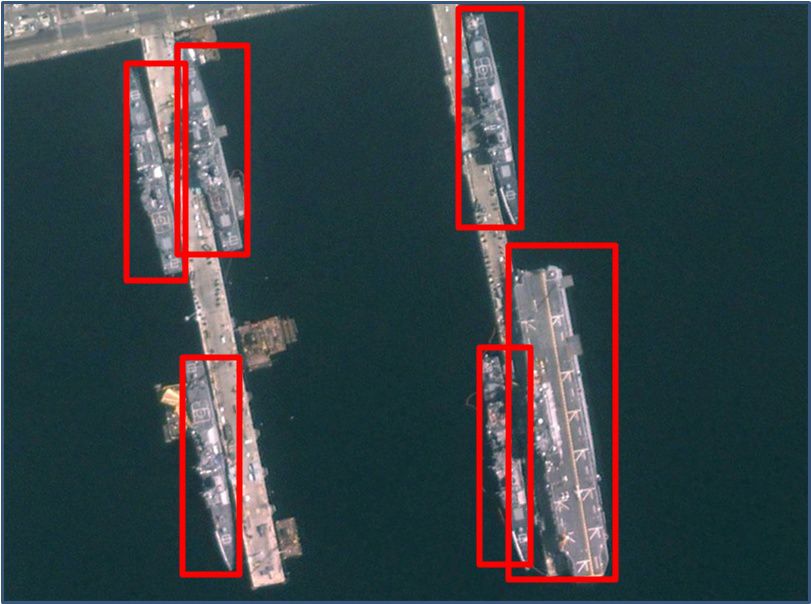}}
%\subfigure{\includegraphics[width=0.16\textwidth]{result_1-6}}
\subfigure{\includegraphics[width=0.19\textwidth]{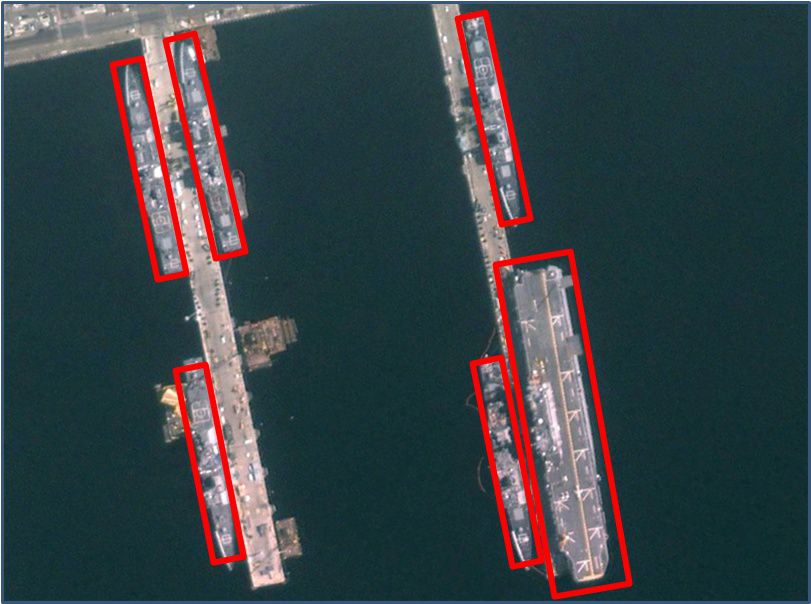}}
\subfigure{\includegraphics[width=0.19\textwidth]{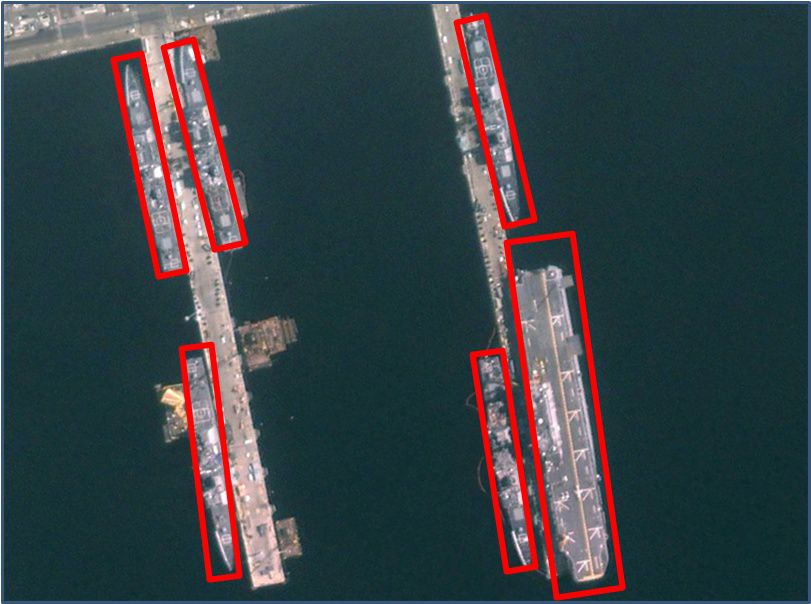}}
\subfigure{\includegraphics[width=0.19\textwidth]{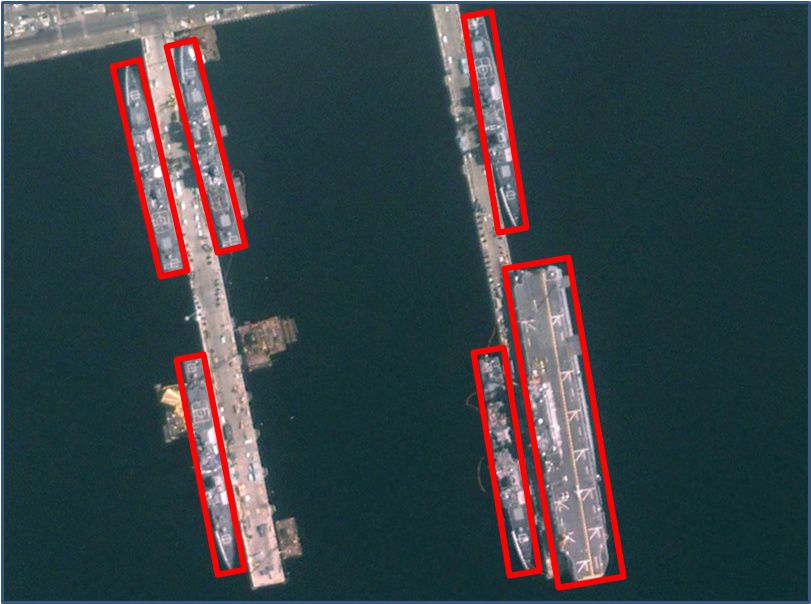}}\\
\vspace{-0.15cm}
\subfigure{\includegraphics[width=0.19\textwidth]{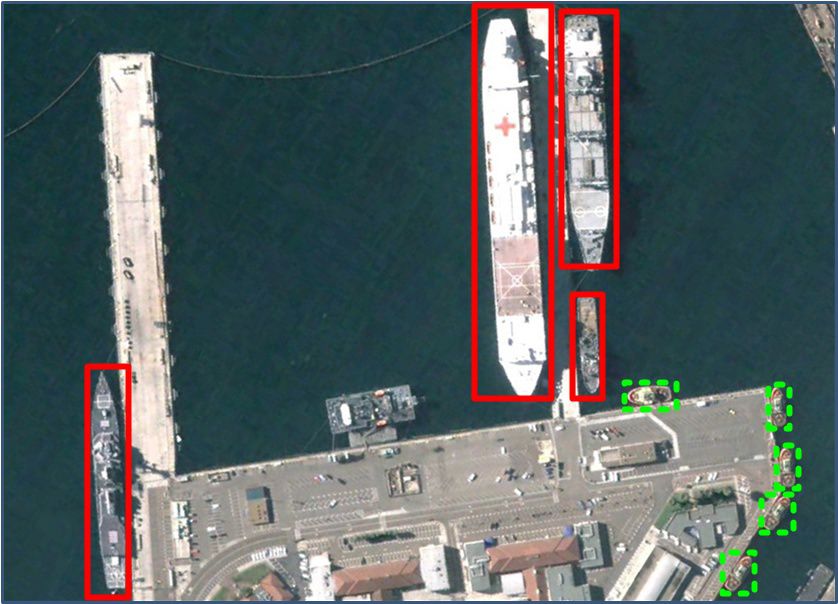}}
\subfigure{\includegraphics[width=0.19\textwidth]{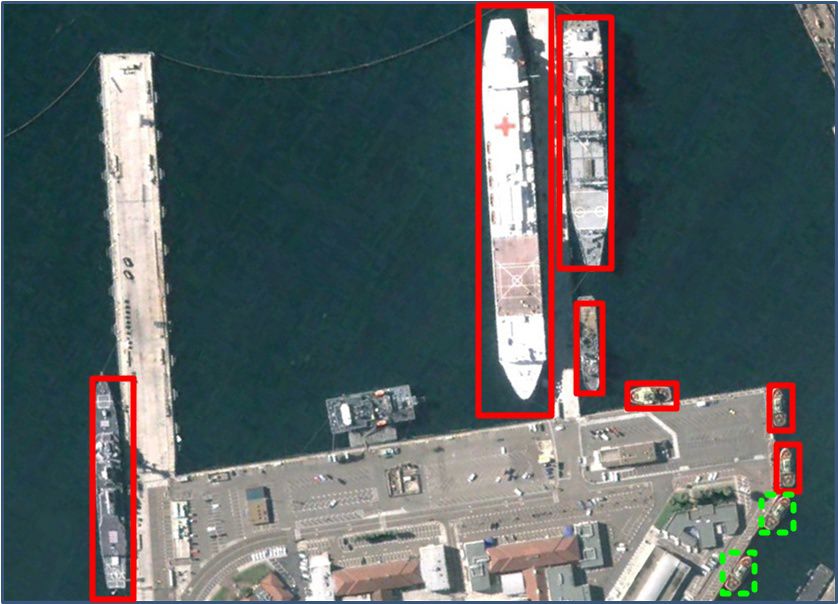}}
%\subfigure{\includegraphics[width=0.16\textwidth]{result_2-6}}
\subfigure{\includegraphics[width=0.19\textwidth]{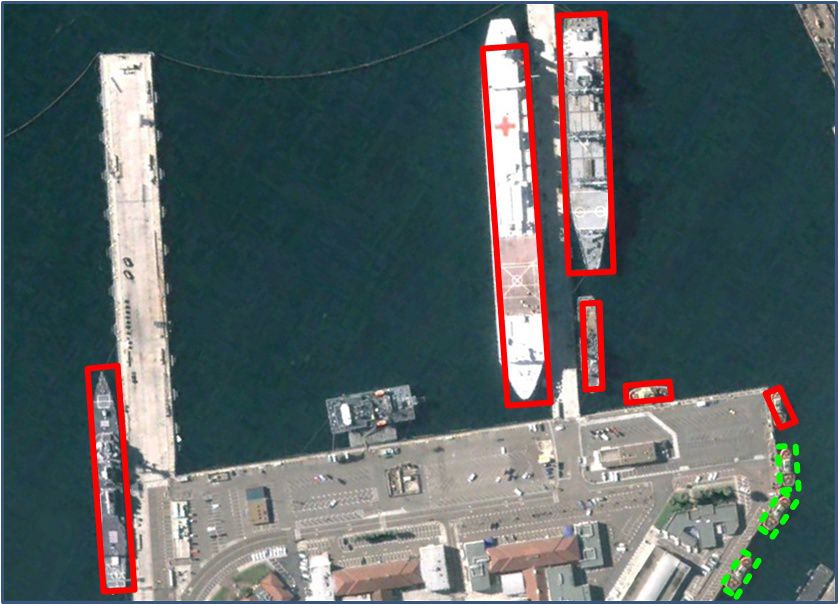}}
\subfigure{\includegraphics[width=0.19\textwidth]{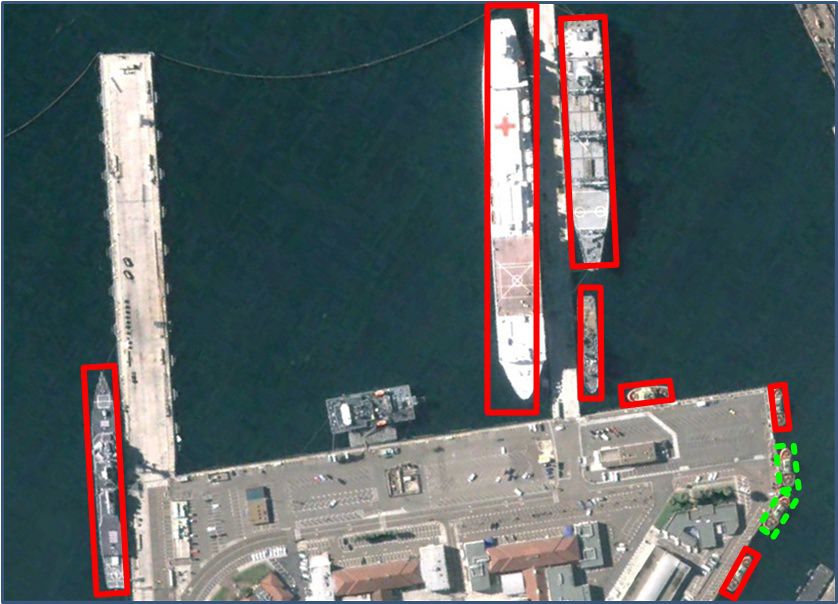}}
\subfigure{\includegraphics[width=0.19\textwidth]{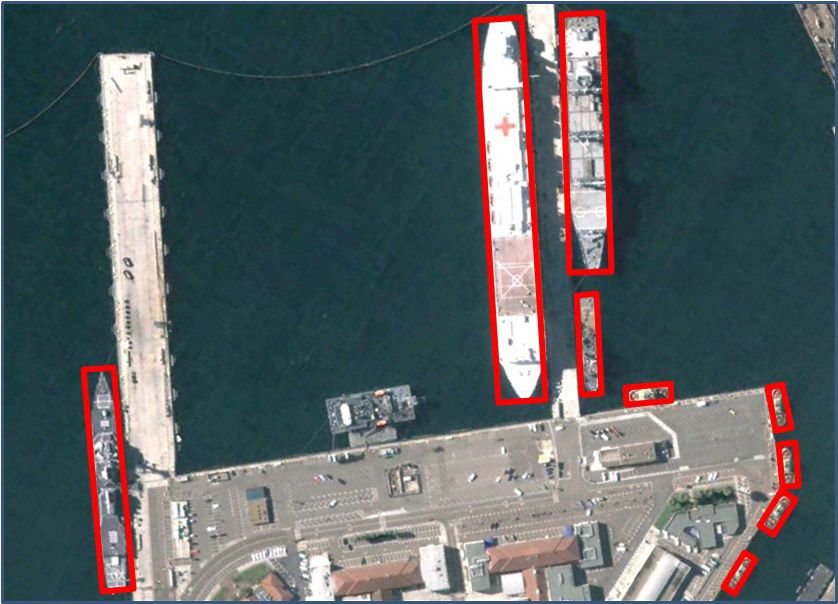}}\\
\vspace{-0.15cm}
\subfigure{\includegraphics[width=0.19\textwidth]{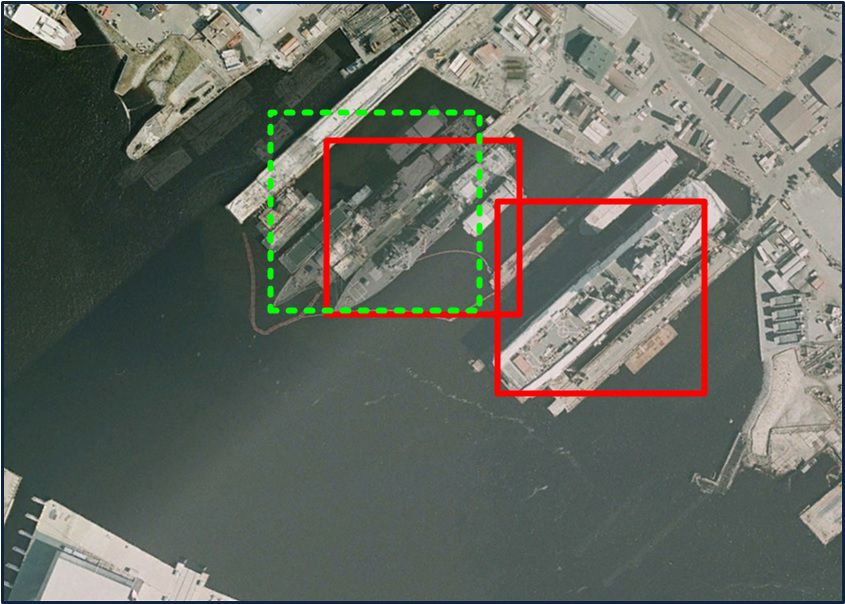}}
\subfigure{\includegraphics[width=0.19\textwidth]{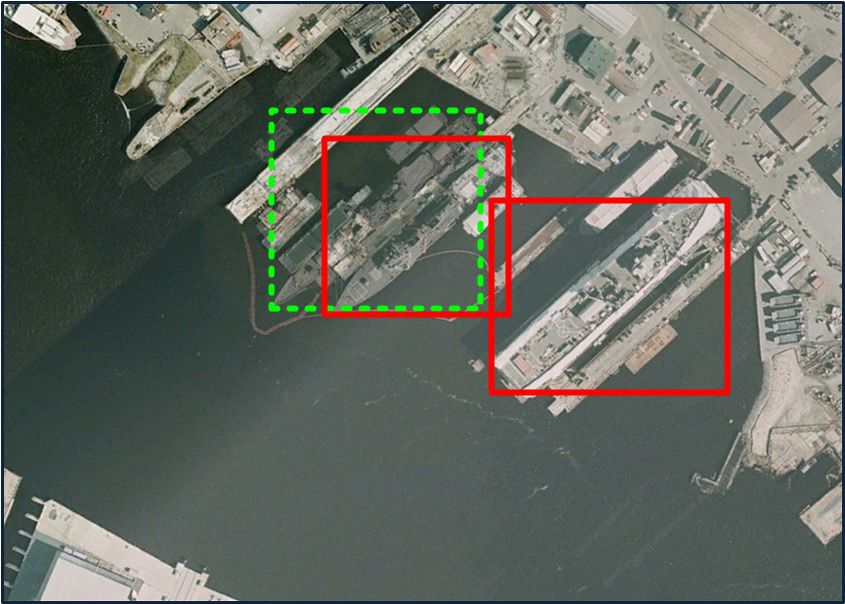}}
%\subfigure{\includegraphics[width=0.16\textwidth]{result_3-6}}
\subfigure{\includegraphics[width=0.19\textwidth]{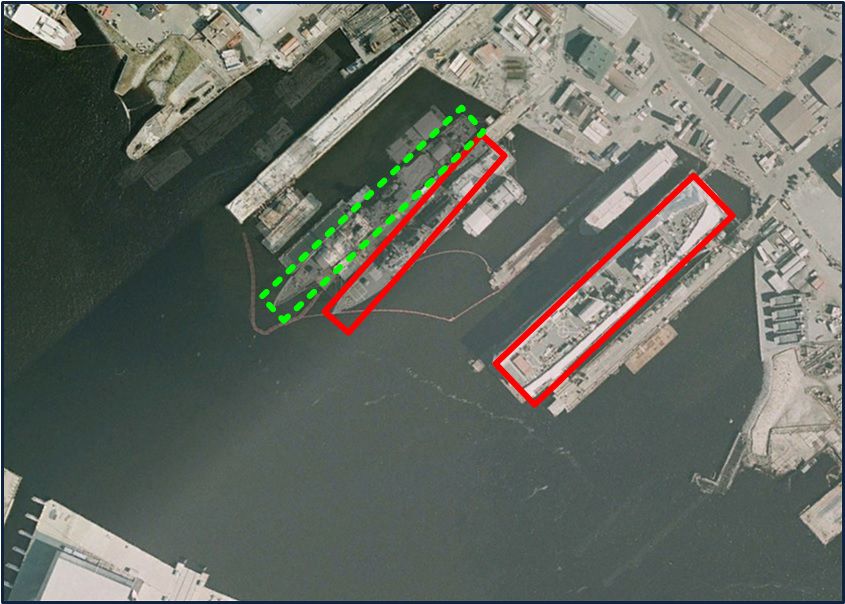}}
\subfigure{\includegraphics[width=0.19\textwidth]{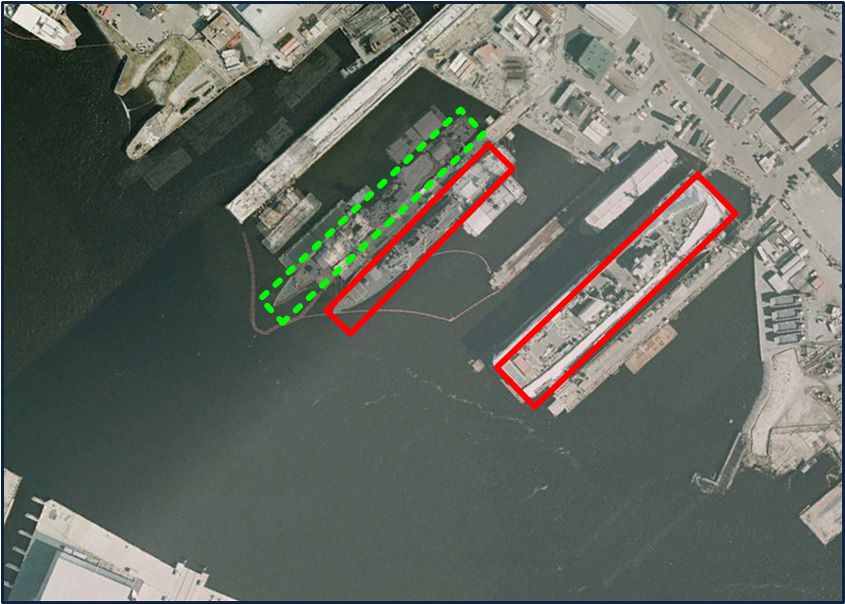}}
\subfigure{\includegraphics[width=0.19\textwidth]{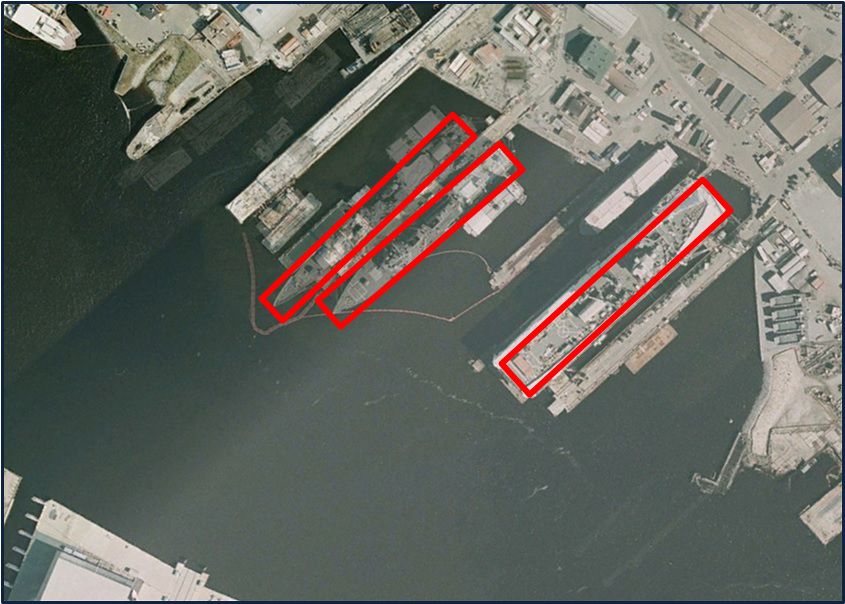}}\\
\vspace{-0.15cm}
\subfigure{\includegraphics[width=0.19\textwidth]{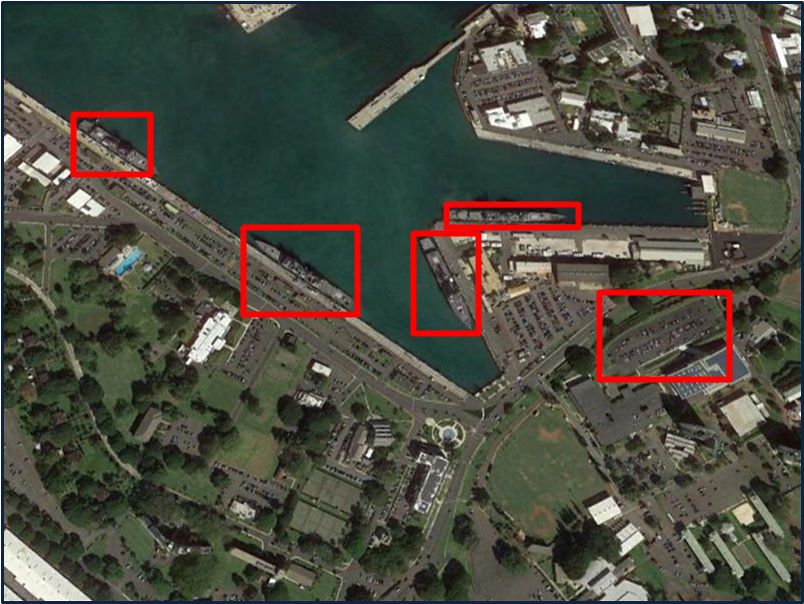}}
\subfigure{\includegraphics[width=0.19\textwidth]{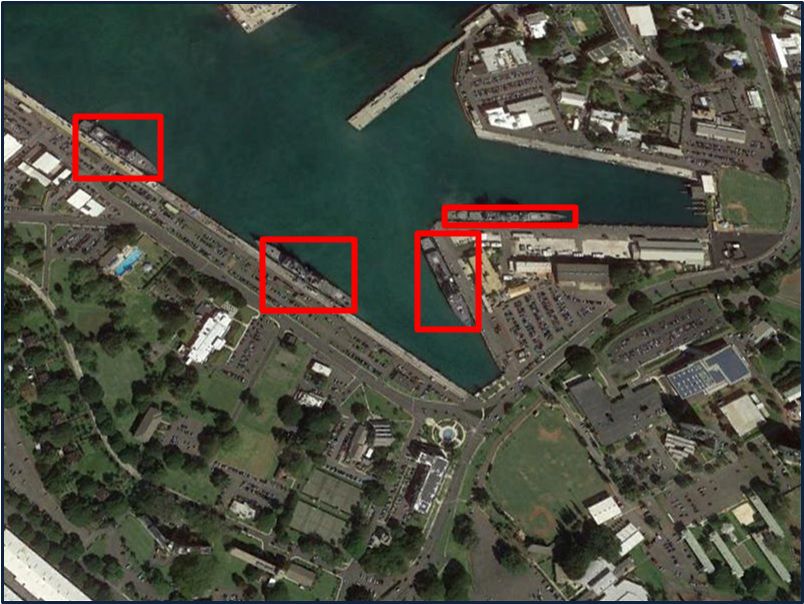}}
%\subfigure{\includegraphics[width=0.16\textwidth]{result_4-6}}
\subfigure{\includegraphics[width=0.19\textwidth]{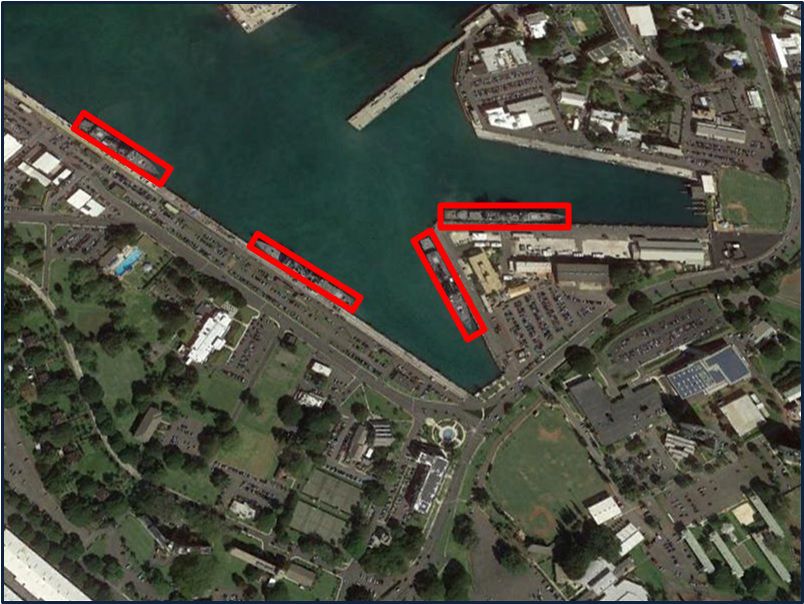}}
\subfigure{\includegraphics[width=0.19\textwidth]{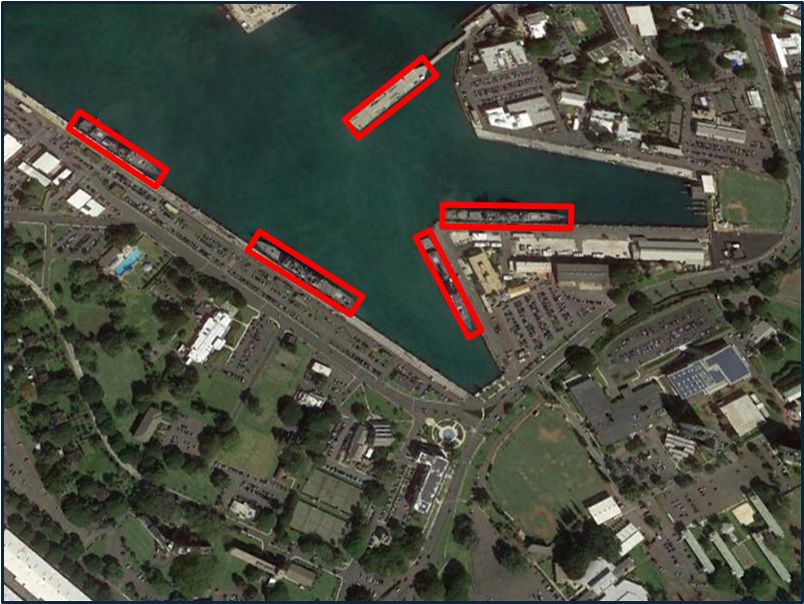}}
\subfigure{\includegraphics[width=0.19\textwidth]{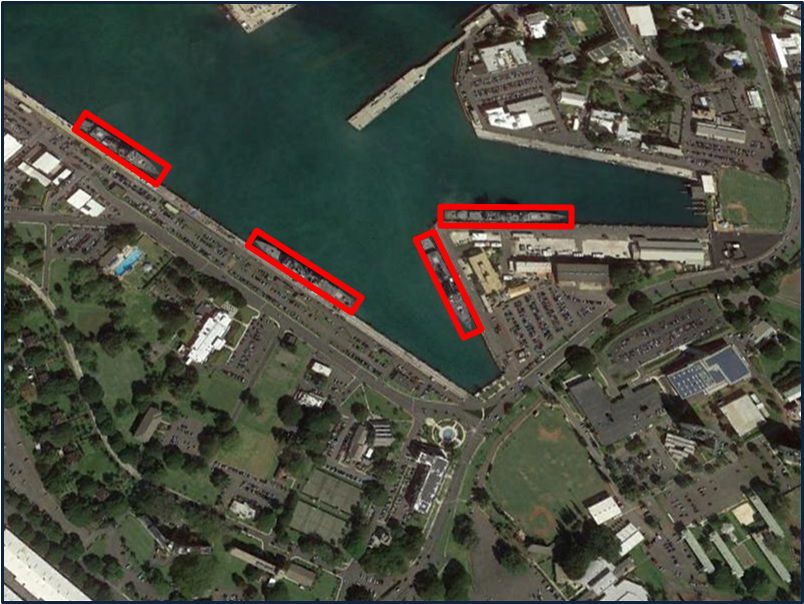}}\\
\vspace{-0.15cm}
\subfigure{\includegraphics[width=0.19\textwidth]{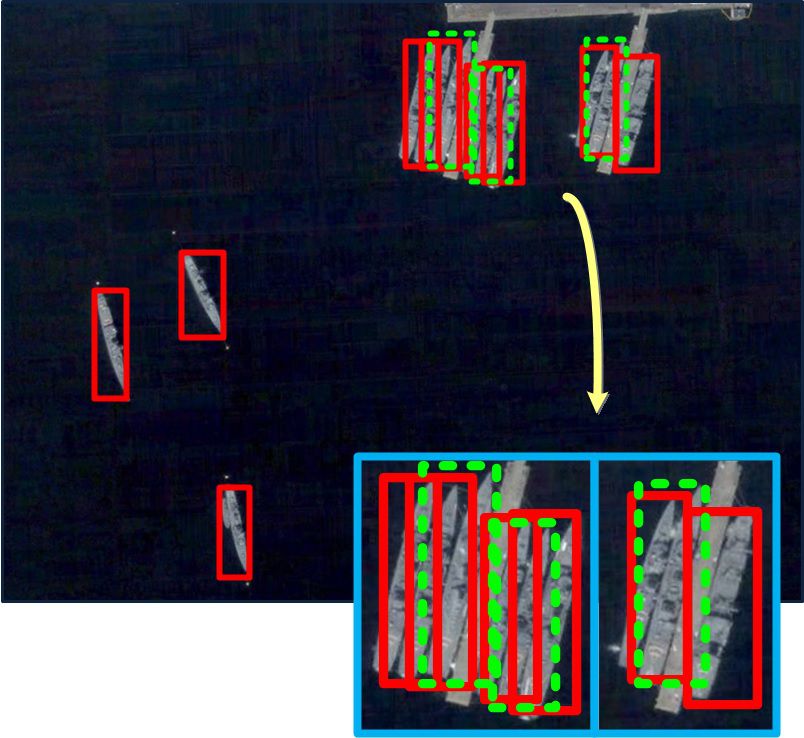}}
\subfigure{\includegraphics[width=0.19\textwidth]{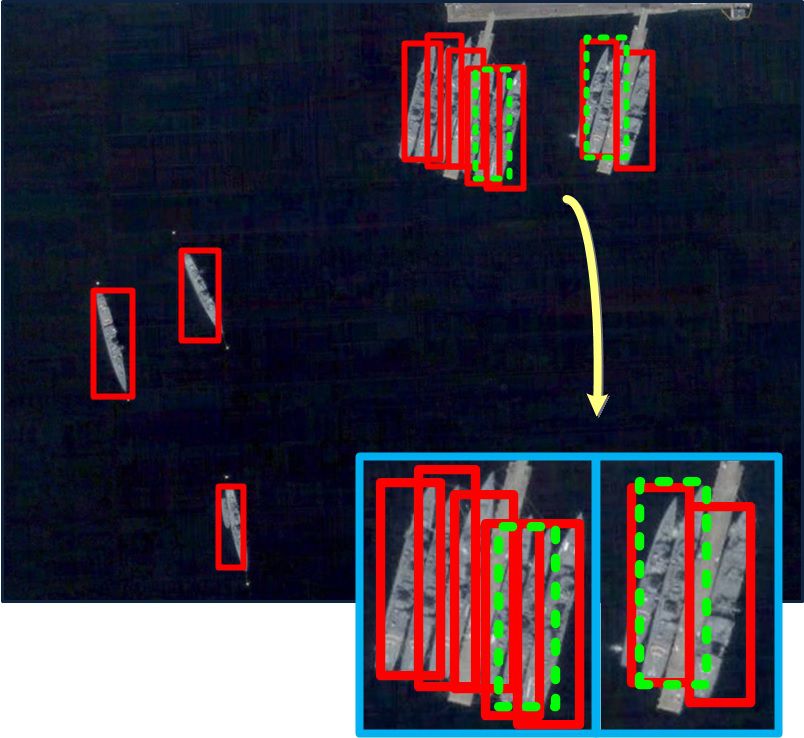}}
%\subfigure{\includegraphics[width=0.16\textwidth]{result_5-6}}
\subfigure{\includegraphics[width=0.19\textwidth]{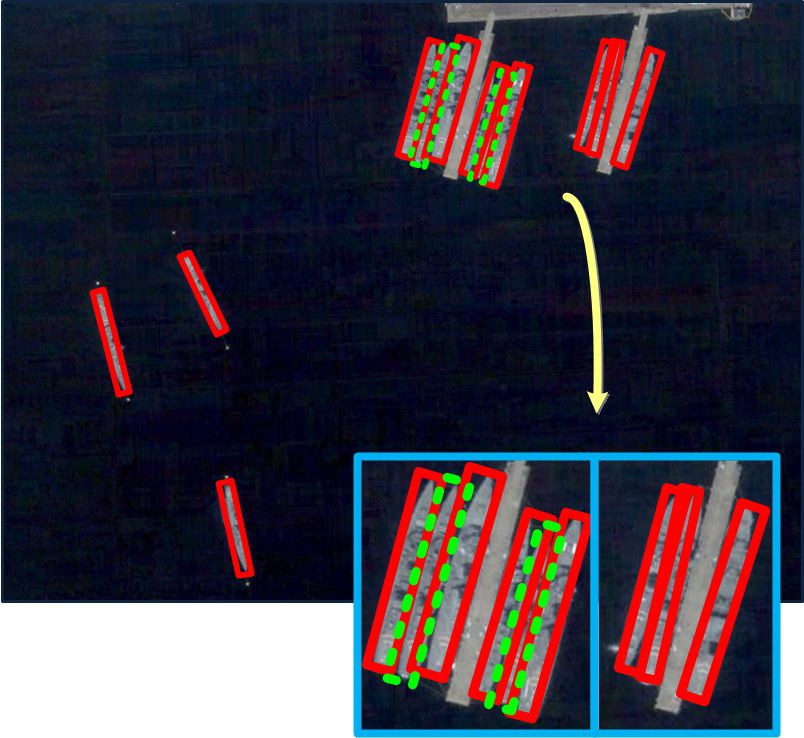}}
\subfigure{\includegraphics[width=0.19\textwidth]{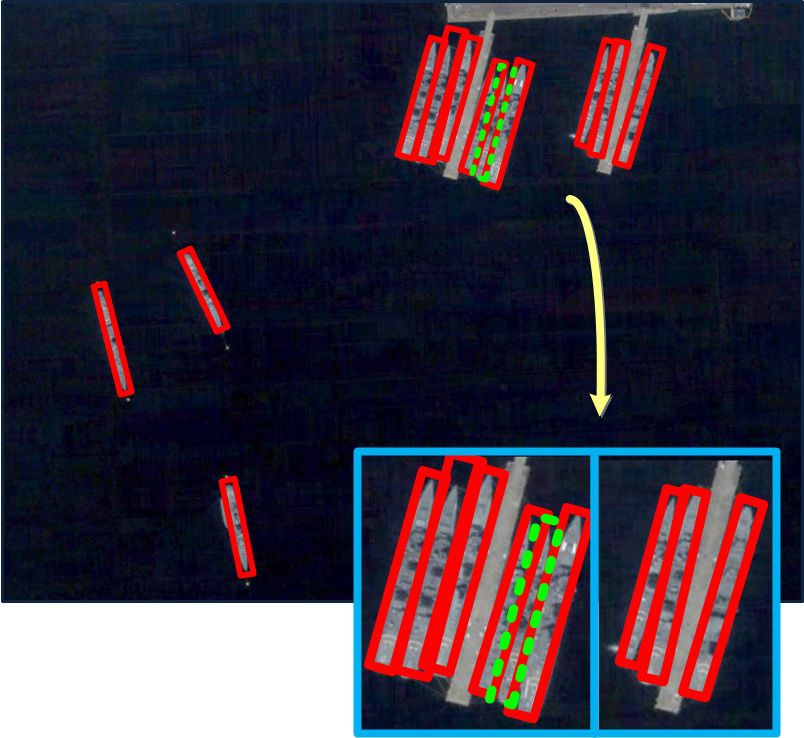}}
\subfigure{\includegraphics[width=0.19\textwidth]{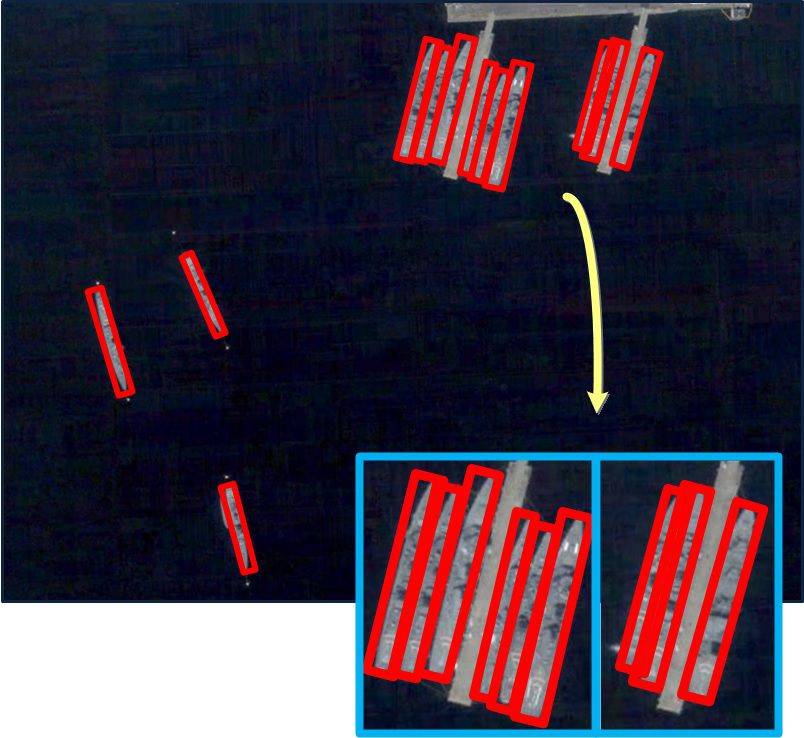}}\\
\vspace{-0.15cm}
\subfigure{\includegraphics[width=0.19\textwidth]{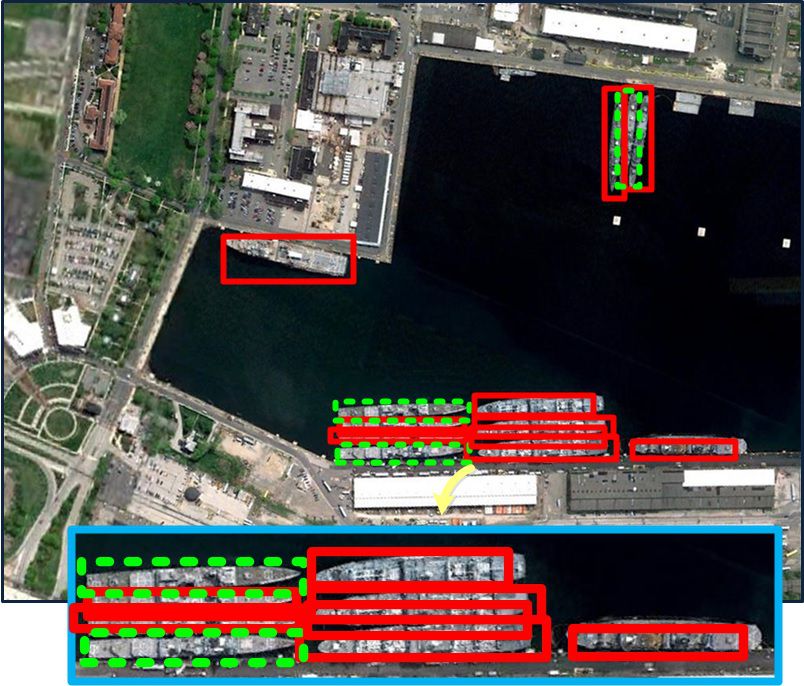}}
\subfigure{\includegraphics[width=0.19\textwidth]{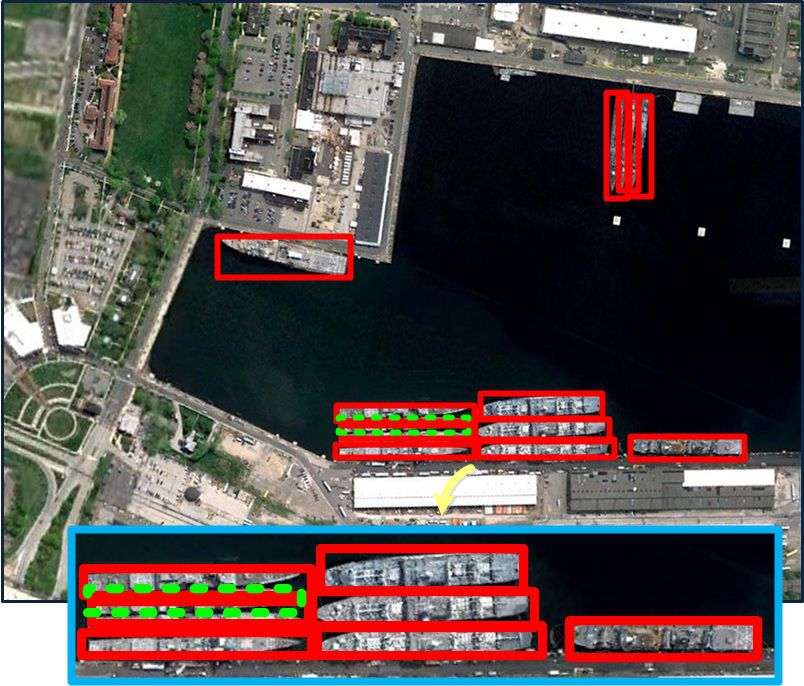}}
%\subfigure{\includegraphics[width=0.16\textwidth]{result_6-6}}
\subfigure{\includegraphics[width=0.19\textwidth]{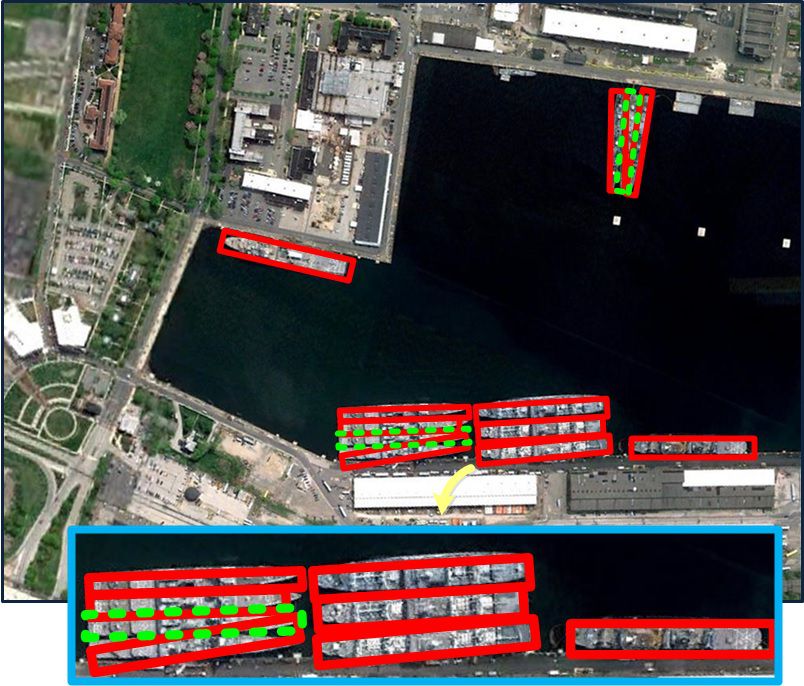}}
\subfigure{\includegraphics[width=0.19\textwidth]{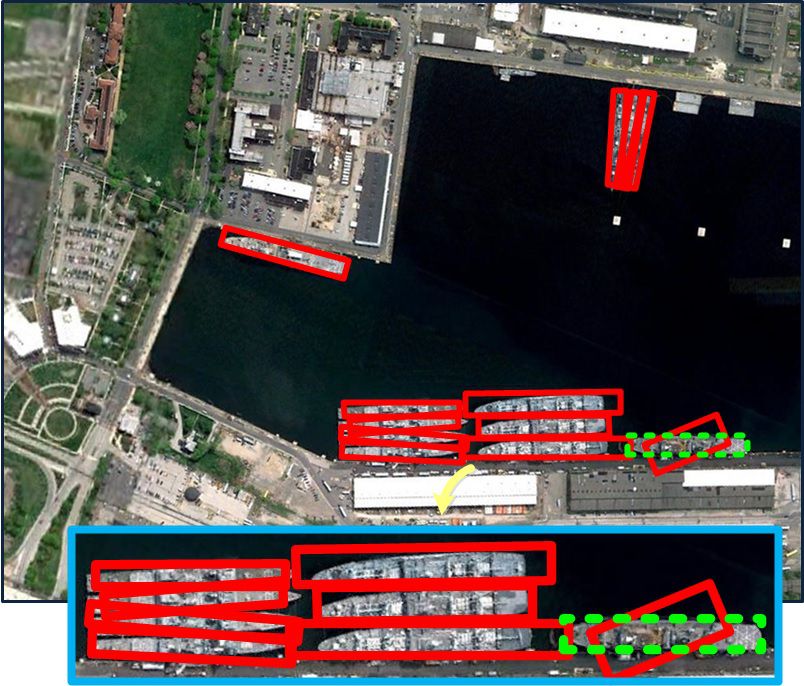}}
\subfigure{\includegraphics[width=0.19\textwidth]{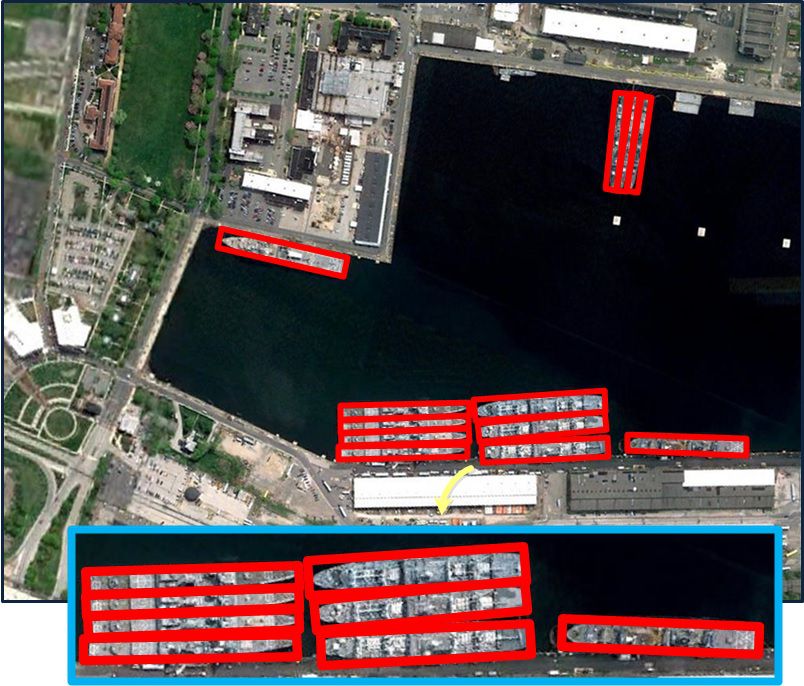}}\\
\vspace{-0.15cm}
\setcounter{subfigure}{0}
\subfigure[Faster R-CNN]{\includegraphics[width=0.19\textwidth]{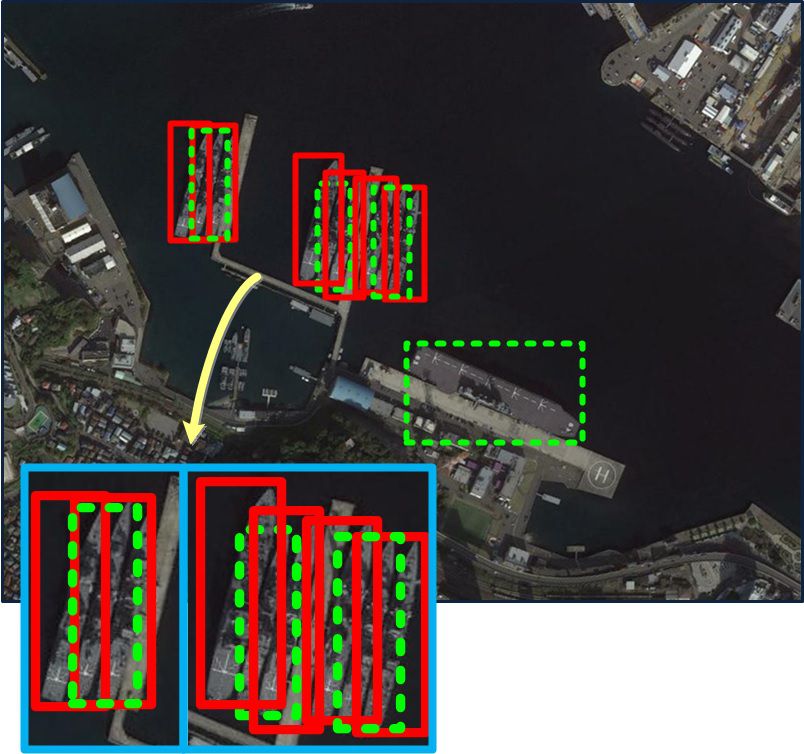}}
\subfigure[YOLOv3]{\includegraphics[width=0.19\textwidth]{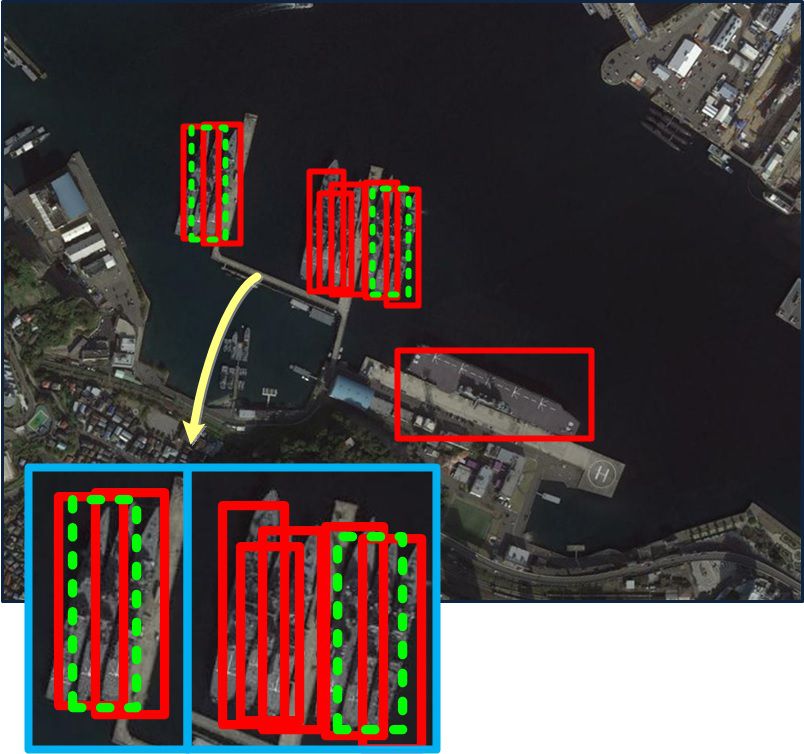}}
%\subfigure[Method \cite{liu2017learning}]{\includegraphics[width=0.16\textwidth]{result_7-6}}
\subfigure[Method \cite{zhang2018toward}]{\includegraphics[width=0.19\textwidth]{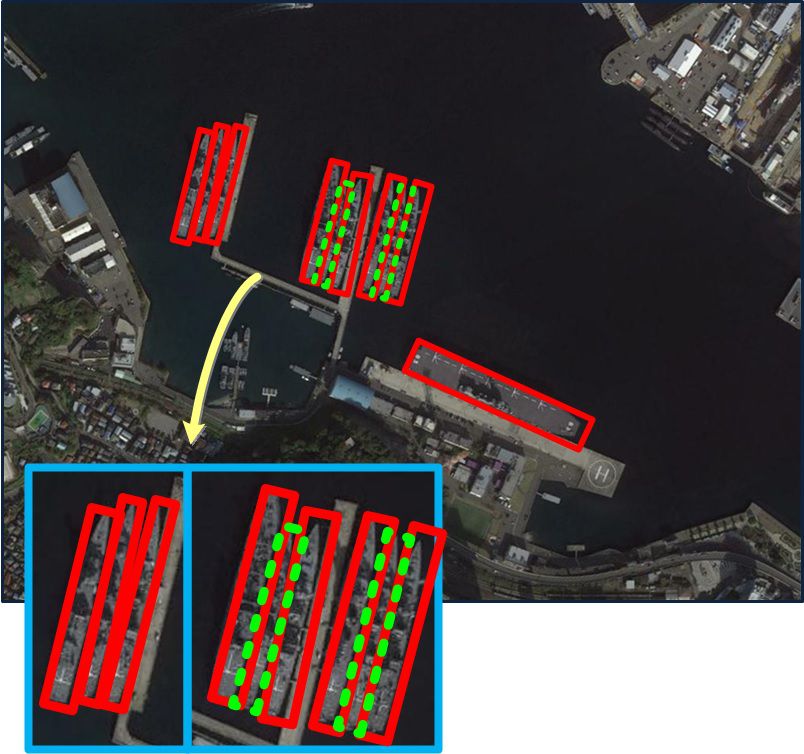}}
\subfigure[R-DFPN]{\includegraphics[width=0.19\textwidth]{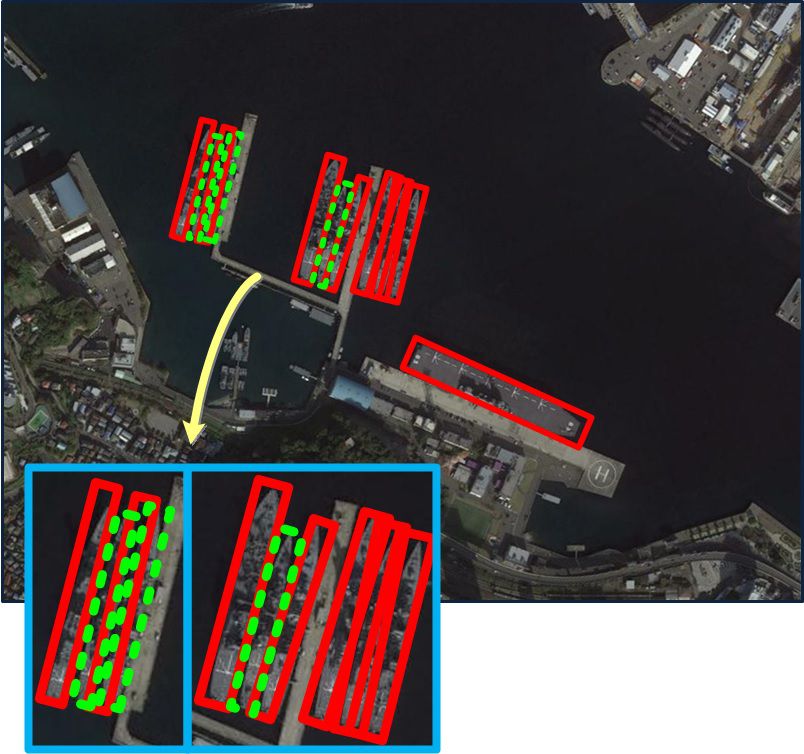}}
\subfigure[Ours]{\includegraphics[width=0.19\textwidth]{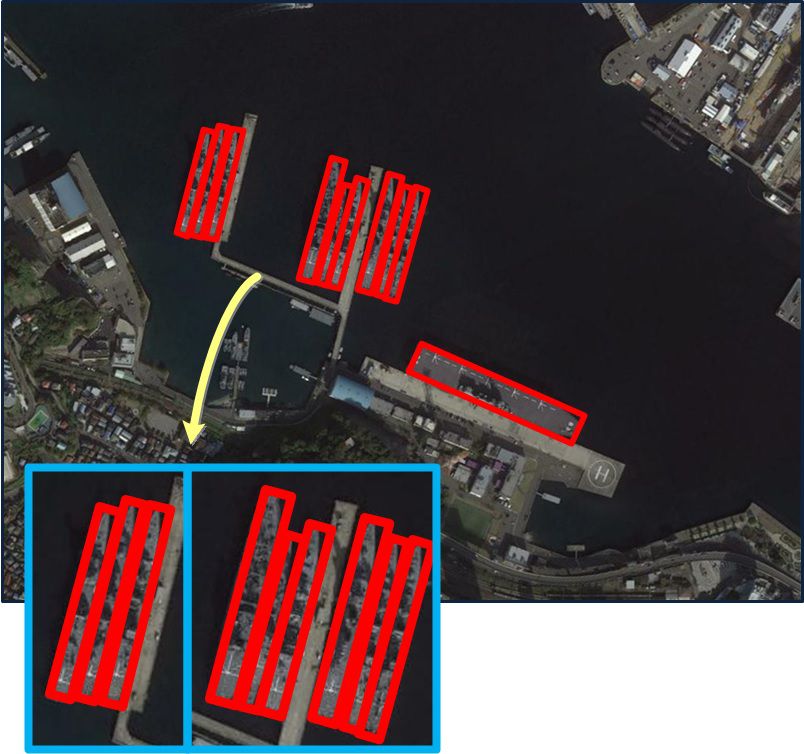}}\\
\caption{Comparison of the ship detection results by different methods.}\label{Fig:Comparison_with_Other_Methods}
\end{figure*}
\subsection{Detection Results and Comparison}
The proposed method is compared with four other representative CNN-based detection methods which are Faster R-CNN \cite{Ren2017Faster}, YOLOv3 \cite{redmon2018yolov3}, method \cite{zhang2018toward} and R-DFPN \cite{yang2018automatic}. Some examples of the detection results by different methods are shown in Fig.~\ref{Fig:Comparison_with_Other_Methods}, where we use the ground-truth bounding boxes plotted in green dashed lines to mark the ships that are unsuccessfully detected in each image. From the first row of Fig.~\ref{Fig:Comparison_with_Other_Methods}, we can see that although all of the methods can successfully detect the ships, our method is able to generate the bounding boxes with much higher accuracy, especially for the lower-left ship in a larger size. Still, in the second row, the other four methods produced less accurate detection results, and even worse, they have failed to detect some of the small ships closely beside the dock. Besides the higher detection accuracy, our method would be with a higher robustness under the interference of complex background (or the other objects with similar appearance). For example, in the third row of Fig.~\ref{Fig:Comparison_with_Other_Methods}, there is a ship surrounded closely by several other objects. While all of the other methods failed to detect this ship, it can be successfully located by our method with a satisfactory accuracy. In the fourth row, we can see that Faster R-CNN and R-DFPN produce false detections on the land and a dock which looks like a ship, respectively.

\renewcommand\arraystretch{1.5}
\begin{table*}[!htp]
\centering
\caption[]{Quantitative comparison results of different methods on the test set}\label{table:AP}
\setlength{\tabcolsep}{6mm}{
\begin{tabular}{lccccc}
\hline
\hline
                &  Faster R-CNN   & YOLOv3   & Method\cite{zhang2018toward}   & R-DFPN  & Ours  \\
\hline
  AP    &  $82.5\%$    &  $87.1\%$   &  $86.6\%$    & $86.8\%$   & $\textbf{89.2\%}$   \\
  Recall    &  $76.4\%$    &  $84.3\%$   &  $83.9\%$    & $85.0\%$   & $\textbf{88.9\%}$   \\
  Precision    &  $84.7\%$    &  $90.5\%$   &  $89.3\%$    & $88.8\%$   & $\textbf{92.0\%}$   \\
\hline
\hline
\end{tabular}}
\end{table*}

\begin{figure*}[!htp]
\centering
\subfigure[]{\includegraphics[width=0.42\textwidth]{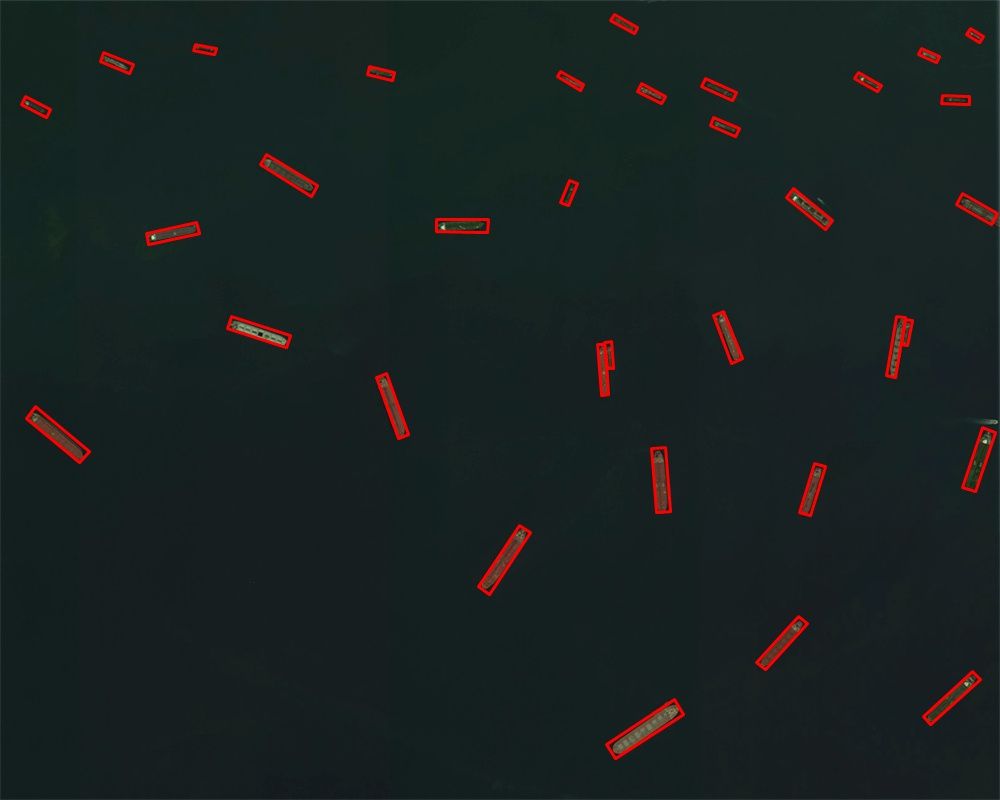}}
\hspace{0.3cm}
\subfigure[]{\includegraphics[width=0.42\textwidth]{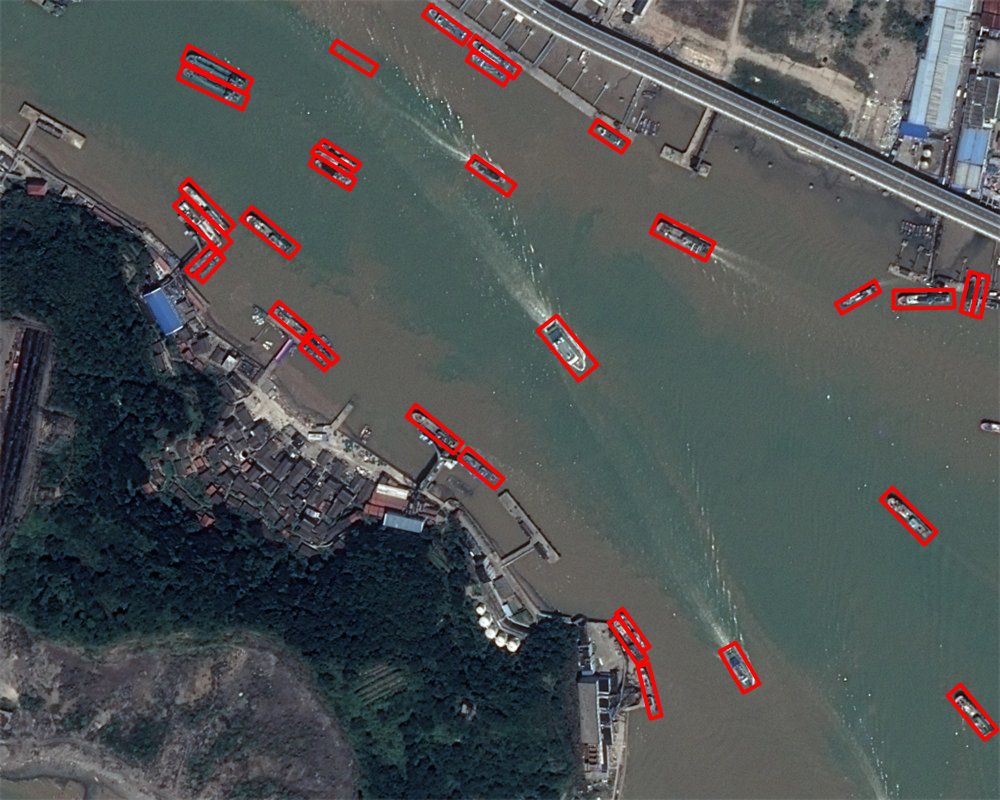}}\\
\subfigure[]{\includegraphics[width=0.42\textwidth]{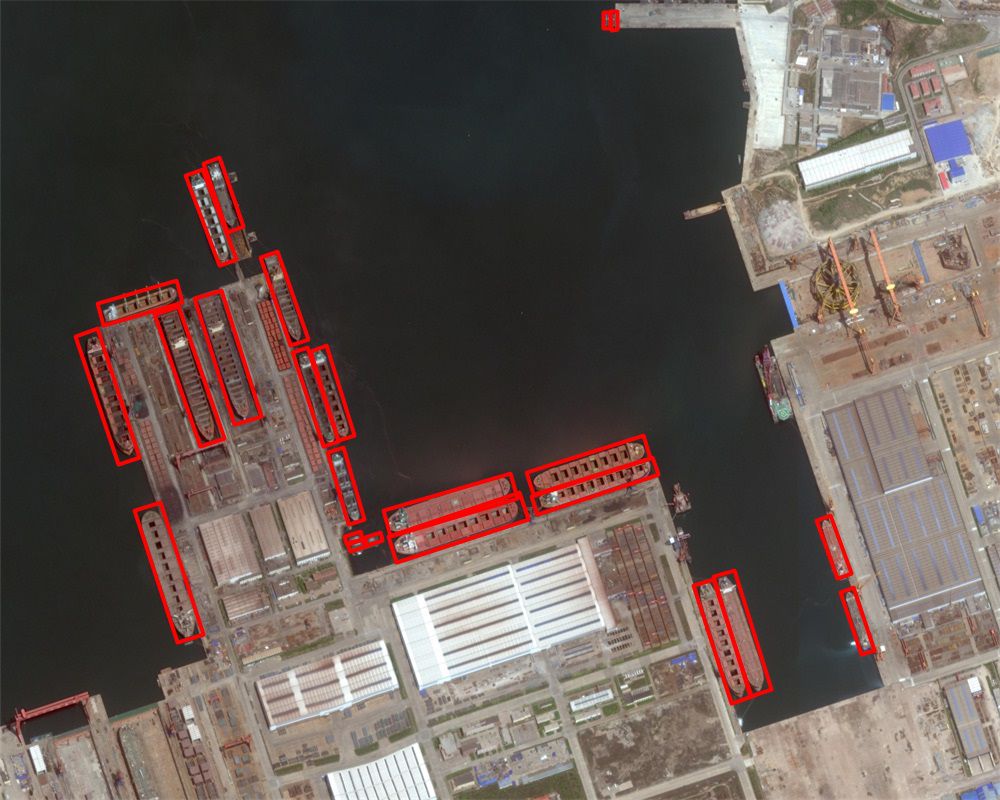}}
\hspace{0.3cm}
\subfigure[]{\includegraphics[width=0.42\textwidth]{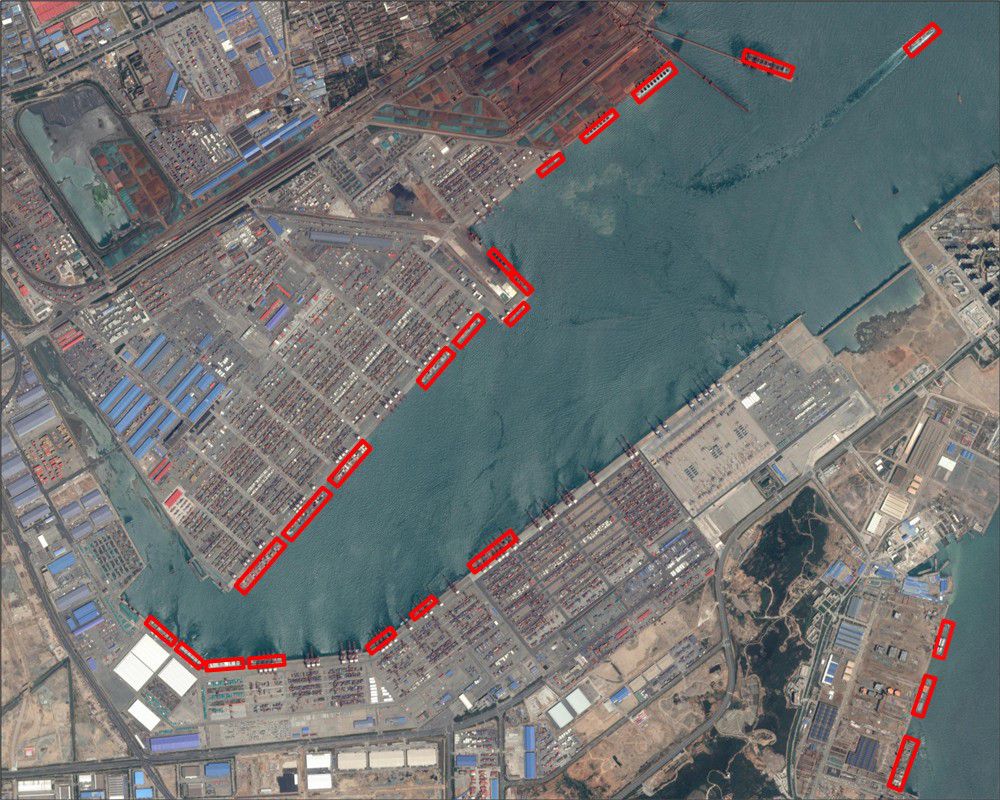}}
\caption{Some examples of detection results obtained by the proposed method.}\label{Fig:big_result}
\end{figure*}
Another challenge is to detect the ships docked densely in the harbor. It would be hard to accurately locate each individual one of the dense ships, especially when they are docked in an inclined orientation or vary in size. The last three rows of Fig.~\ref{Fig:Comparison_with_Other_Methods} provides some examples. From the enlarged image patches, we can see that Faster R-CNN and YOLOv3, which detect the ships with horizontal bounding boxes, are prone to producing more inaccurate or miss detection results. The method\cite{zhang2018toward} and R-DFPN perform better in accurately locating the individual ships by using rotated bounding boxes based on the orientation prediction. However, there still exist some miss detections when the narrow ships docked densely and very closely to each other. Moreover, they can not guarantee to provide the correct orientation of the bounding box (see the last but one row of Fig.~\ref{Fig:Comparison_with_Other_Methods}), due to the difficulty to reliably predict the orientation together with other variables in one regression process. By contrast, our method is able to obtain much better detection results in the above cases. More detection results obtained by our method can be seen in Fig.~\ref{Fig:big_result}.

\begin{figure}[!htp]
\centering
\subfigure{\includegraphics[width=0.45\textwidth]{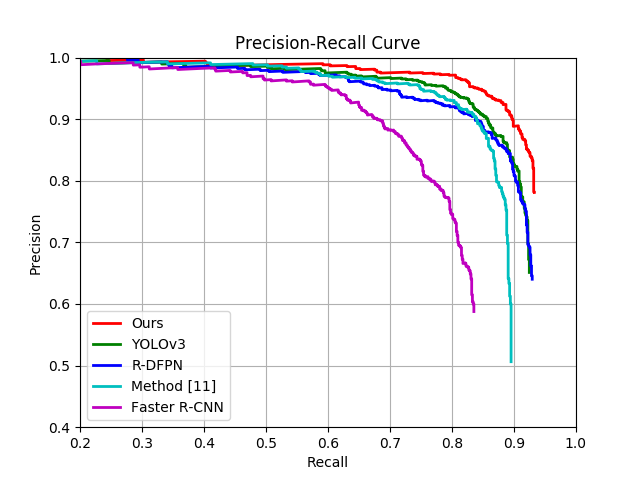}}
\caption{The Precision-Recall curves of different detection methods.}\label{Fig:PR_Curve}
\end{figure}

Quantitative comparisons on the test set are provided in Table~\ref{table:AP} and Fig.~\ref{Fig:PR_Curve}. From Table~\ref{table:AP}, it can be seen that our method obtains the highest scores on all the indexes of precision, recall and AP. The method of R-DFPN achieves slightly higher AP compared with the method\cite{zhang2018toward}, while Faster R-CNN gets the worst performance. Although YOLOv3 achieves higher AP than R-DFPN, its recall is lower. This is probably because of the more miss detections obtained by YOLOv3 for the inclined dense ships.

Fig.~\ref{Fig:PR_Curve} presents a more comprehensive comparison of the above methods with the precision-recall curves. It can be seen that our method outperforms the others remarkably at the majority of points on the curve. With the elaborately-designed more powerful backbone network and the undergoing of several versions of upgradation in algorithm, the YOLOv3 can now have very good performance to detect various objects, and has become a highly competitive method in many object-detection tasks\cite{zhao2019m2det, braun2019eurocity, benjdira2019car, ghiasi2019fpn}. In Fig.~\ref{Fig:PR_Curve}, we can see that overall YOLOv3 achieves the second-best performance for ship detection, except that it would obtain lower recall than R-DFPN when the precision drops. The fact that R-DFPN as well as the method\cite{zhang2018toward} shows the inferior performance than YOLOv3, to some extent reveals the difficulty of detecting ships via rotated bounding box. That is, the unknown variable of orientation additionally introduced into the system would increase the complexity, and if not properly addressed it would in turn affect the performance of algorithm. We can see that our method has well resolved the relevant problems and achieves a boost in performance, though a more powerful backbone network (as well as the other sophisticated techniques) is not used like YOLOv3.

From Fig.~\ref{Fig:PR_Curve}, we can see that R-DFPN and the method\cite{zhang2018toward} overall achieve comparable performance. The difference is that R-DFPN would be with a higher recall but a lower precision, and converse for the method\cite{zhang2018toward}. In fact, the two compared methods can be recognized as two opposite extreme cases in the feature employment for ship detection. The method\cite{zhang2018toward} only employs a single-level raw features from the backbone, while R-DFPN uses the complicated multilevel features from densely-connected feature pyramid. It is generally believed that using more sophisticated multilevel features would be helpful to improve the detection performance. However, this may also increase the risk of over-fitting for the ship targets due to the increased complexity of network, and hinder the extraction of intrinsic features. In this case, it would generate more false detections, which might be one of the causes to make R-DFPN have the lower precision. By contrast, our method can alleviate this problem via a novel compact representation of multilevel features.
\section{Conclusion} \label{section:conclusion}
In this paper, we propose a novel CNN-based method to detect the ships in high-resolution optical remote sensing images via rotated bounding box. The proposed method first  generates high-quality rotated proposals with a dual-branch regression network. Compared with other methods which handle all unknown variables together with shared features in one regression process, the dual-branch regression network predicts the orientation and other variables independently with particular CNN-features that are more suitable for corresponding regression task. Next, a multilevel adaptive pooling method is proposed to alleviate the uneven sampling problem caused by typical pooling methods, and meanwhile incorporate multilevel features via a spatially-variant pooling operation. This novel approach creates a compact feature representation enabling to fully take advantages of multilevel features to improve the detection performance. Finally, detailed ablation study is performed on the proposed techniques, and the superiority of the proposed detection method is verified by comparing it with some representative CNN-based detection methods, from which some useful insights regarding a proper use of multilevel CNN-features for ship detection are provided as well.

\ifCLASSOPTIONcaptionsoff
  \newpage
\fi

%\bibliography{mybibfile}
%\bibliographystyle{IEEEtran}


%\begin{IEEEbiography}{Linhao Li}
%\end{IEEEbiography}

%\begin{IEEEbiographynophoto}{Zhiqiang Zhou}

%\newpage

%\begin{IEEEbiographynophoto}{Bo Wang}


%\vfill


\end{document}